\documentclass[10pt,twocolumn,letterpaper]{article}

\usepackage[pagenumbers]{cvpr} 

\usepackage{graphicx}
\usepackage{amssymb}
\usepackage[bb=boondox]{mathalfa}

\usepackage{booktabs}

\usepackage{caption}
\usepackage{adjustbox}
\usepackage{algorithmic}
\usepackage{amsmath}
\usepackage[ruled,vlined]{algorithm2e}

\usepackage{xr}
\makeatletter
\newcommand*{\addFileDependency}[1]{
  \typeout{(#1)}
  \@addtofilelist{#1}
  \IfFileExists{#1}{}{\typeout{No file #1.}}
}
\makeatother

\usepackage[pagebackref,breaklinks,colorlinks]{hyperref}

\usepackage[capitalize]{cleveref}
\crefname{section}{Sec.}{Secs.}
\Crefname{section}{Section}{Sections}
\Crefname{table}{Table}{Tables}
\crefname{table}{Tab.}{Tabs.}

\usepackage{multirow}

\DeclareMathOperator*{\argmax}{arg\,max}

\def\cvprPaperID{10484} 
\def\confName{CVPR}
\def\confYear{2022}

\begin{document}

\title{Towards Robust and Reproducible Active Learning using Neural Networks}

\author{Prateek Munjal \thanks{G42 Healthcare, Abu Dhabi, UAE}
\and
Nasir Hayat \thanks{NYUAD, UAE}
\and
Munawar Hayat \thanks{Monash University, Australia}
\and
Jamshid Sourati \thanks{University of Chicago, Chicago, IL, USA}
\and
Shadab Khan \footnotemark[1]
}




\newcommand\blfootnote[1]{%
  \begingroup
  \renewcommand\thefootnote{}\footnote{#1}%
  \addtocounter{footnote}{-1}%
  \endgroup
}

\maketitle
\blfootnote{Correspondence to: Shadab Khan
\textless skhan.shadab@gmail.com\textgreater}
\begin{abstract}
    Active learning (AL) is a promising ML paradigm that has the potential to parse through large unlabeled data and help reduce annotation cost in domains where labeling data can be prohibitive. Recently proposed neural network based AL methods use different heuristics to accomplish this goal. In this study, we demonstrate that under identical experimental settings, different types of AL algorithms (uncertainty based, diversity based, and committee based) produce an inconsistent gain over random sampling baseline. Through a variety of experiments, controlling for sources of stochasticity, we show that variance in performance metrics achieved by AL algorithms can lead to results that are not consistent with the previously reported results. We also found that under strong regularization, AL methods show marginal or no advantage over the random sampling baseline under a variety of experimental conditions. Finally, we conclude with a set of recommendations on how to assess the results using a new AL algorithm to ensure results are reproducible and robust under changes in experimental conditions. We share our codes to facilitate AL evaluations. We believe our findings and recommendations will help advance reproducible research in AL using neural networks. We open source our code at 
    \href{https://github.com/PrateekMunjal/TorchAL}{\textcolor{blue}{https://github.com/PrateekMunjal/TorchAL}}
\end{abstract}

\hspace{-6.4mm}\textbf{Abbreviations}: Active Learning (AL), Random Sampling Baseline (RSB), Query-by-Committee (QBC), Variational Adversarial Active Learning (VAAL), Uncertainty-based sampling (UC), Deep Bayesian Active Learning (DBAL), Bayesian Active Learning by Disagreement (BALD), Random Augmentation (RA), Stochastic Weighted Averaging (SWA), Shake-Shake regularization (SS).

\hspace{-6.4mm}\textbf{Conventions}: Models that are regularized using either one or combination of RA, SWA, and SS have been identified with a suffix \textit{-SR} added to the abbreviation, SR signifies `strong regularization'. Models without such a suffix are also regularized, but with standard methods such as weight decay and data augmentation using random flip and horizontal crop.

\section{Introduction}
\label{sec:intro}

Active learning (AL) is a machine learning paradigm that promises to help reduce the burden of data annotation by intelligently selecting a subset of informative samples from a large pool of unlabeled data that. In AL, a model trained with a small amount of labeled seed data is used to parse through the unlabeled data to select the subset that should be sent to an oracle (annotator). To select such a subset, AL methods rely on exploiting the learned latent-space, model uncertainty, or other heuristics. The promise of reducing annotation cost has brought a surge of interest in AL research \cite{VAAL_sinha2019variational, coreset_sener2018active, Ensembles_Beluch2018ThePO, DBAL_gal2017deep,BatchBald_DBLP:journals/corr/abs-1906-08158, BGALD_DBLP:journals/corr/abs-1904-11643,yoo2019learning_loss_for_AL,coregcn_caramalau2021sequential, tavaal_kim2021task} and with it, a few outstanding issues.
\textbf{First}, The results reported for Random sampling baseline, RSB vary significantly between studies. For example, using $20\%$ labeled data of CIFAR10, the difference between RSB performance reported by \cite{yoo2019learning_loss_for_AL} and \cite{BGALD_DBLP:journals/corr/abs-1904-11643} is $\approx 13\%$ under identical settings. 
\textbf{Second}, 
The results reported for the same AL method can vary across studies: using VGG16 \cite{vgg16_simonyan2014very} on CIFAR100 \cite{cifar10_dataset_krizhevsky2009learning} with $40\%$ labeled data, Coreset \cite{coreset_sener2018active} reported $\approx 55\%$ classification accuracy whereas VAAL ~\cite{VAAL_sinha2019variational} reported $47.01\%$ using the method reported in \cite{coreset_sener2018active}. 
\textbf{Third}, Recent AL studies have been inconsistent with each other. For example, \cite{coreset_sener2018active} and \cite{Adv_AL_DBLP:journals/corr/abs-1802-09841} state that diversity-based AL methods consistently outperform uncertainty-based methods, which were found to be worse than the RSB. In contrast, recent developments in uncertainty based studies \cite{yoo2019learning_loss_for_AL} suggest otherwise.



\noindent In addition to these issues, results using a new AL method are often reported on simplistic experimental conditions - (i) regularization is not sufficiently explored beyond the usual methods (e.g. weight decay), (ii) with increasing AL iterations, the training data distribution changes, however, the training hyper-parameters are fixed in advance. Such issues in the AL results has spurred a recent interest in benchmarking of AL methods and recent NLP and computer vision studies have raised a number of interesting questions \cite{inspire1_DBLP:journals/corr/abs-1807-04801, inspire2_prabhu2019sampling,mittal2019parting}. With the goal of improving the reproducibility and robustness of AL methods, in this study we evaluate the performance of these methods for image classification task.

\noindent \textbf{Contributions}: Through a comprehensive set of experiments performed using consistent settings under a common code base (PyTorch-based\footnote{\url{https://github.com/PrateekMunjal/TorchAL}})
we compare different AL methods including state-of-the-art diversity-based, uncertainty-based, and committee-based methods \cite{VAAL_sinha2019variational, coreset_sener2018active, Ensembles_Beluch2018ThePO, DBAL_gal2017deep} and a well-tuned RSB. We demonstrate that:
\textbf{1)} With strong regularization and hyper-parameters tuned using AutoML, 
RSB performs comparably to AL methods in contrast to the previously reported results in the literature.
    \textbf{2)} No AL method consistently outperforms other approaches, and conclusions can change with different experimental settings (\eg using a different architecture for the classifier or with different number of AL iterations).
    \textbf{3)} The difference in performance between the AL methods and the RSB is much smaller than reported in the literature. 
    \textbf{4)} With a strongly-regularized model, the variance in accuracy achieved using AL methods is substantially lower across consistent repeated training runs, suggesting that such a training regime is unlikely to effect misleading results in AL experiments.
    \textbf{5)} Finally, we provide a set of guidelines on experimental evaluation of a new AL method.

\section{Pool Based Active Learning Methods}

Contemporary pool-based AL methods can be broadly classified into: \textbf{(i)} uncertainty based \cite{VAAL_sinha2019variational,DBAL_gal2017deep,BatchBald_DBLP:journals/corr/abs-1906-08158}, \textbf{(ii)} diversity based \cite{coreset_sener2018active, Adv_AL_DBLP:journals/corr/abs-1802-09841}, and \textbf{(iii)} committee based \cite{Ensembles_Beluch2018ThePO}. AL methods also differ in other aspects, for example, some AL methods use the task model (e.g. model trained for image classification) within their sampling function \cite{DBAL_gal2017deep, coreset_sener2018active}, where as others use different models for task and sampling functions \cite{VAAL_sinha2019variational , Ensembles_Beluch2018ThePO}.
These methods are discussed in detail next.

\textbf{Notations}: Starting with an initial set of labeled data $L_{0}^{0}$=$\{(x_i, y_i)\}_{i=1}^{N_L}$ and a large pool of unlabeled data $U_{0}^{0}$=$\{ x_i\}_{i=1}^{N_U}$, pool-based AL methods train a model $\Phi_0$. A sampling function $\Psi(L_{0}^{0}, U_{0}^{0}, \Phi_0)$ then evaluates $x_i \in U_0$, and selects $k$ (budget size) \textbf{samples} to be labeled by an oracle. The selected samples with \textbf{oracle-annotated} labels are then added to $L_{0}^{0}$, resulting in an extended $L_{0}^{1}$ labeled set, which is then used to \textbf{retrain} $\Phi$. This cycle of \textbf{sample-annotate-train} is repeated until the sampling budget is exhausted or a satisficing metric is achieved. AL sampling functions evaluated in this study are outlined next. 

\subsection{Model Uncertainty on Output (UC)}
\cite{uncertainty_lewis1994sequential} ranks the unlabeled data, $x_i \in U$ in a descending order based on their scores given by $\max_{j}\Phi(x_i); j\in\{1\dots C\}$
where $C$ is the number of classes, and chose the top $k$ samples. Typically this approach focuses on the samples in $U$ for which the softmax classifier is least confident. Recently, Huang et al.\cite{tod_huang2021semi} proposed to measure the uncertainty by measuring the output discrepancies from the model trained at different AL cycles.

\subsection{Deep Bayesian Active Learning (DBAL)}

Gal et al. \cite{DBAL_gal2017deep} train the model $\Phi$ with dropout layers and use monte carlo dropout to approximate the sampling from posterior.  
For our experiments, we used the two most reported acquisitions \ie  max entropy and Bayesian Active Learning by Disagreement (BALD). The max entropy method selects the top $k$ data points having maximum entropy as $\argmax_i \mathbb{H} [P(\textbf{y}|x_i)]; \forall x_i \in U_0$
where the posterior is given by, 
$P(\textbf{y}|x_i) = \sum_{j=1}^{T}\frac{1}{T} P(\textbf{y}|x_i, \phi_{j})$.
Here $T$ denotes number of forward passes through the model, $\Phi$. 
BALD selects the top $k$ samples that increase the information gain over the model parameters \ie $
\argmax_i \mathbb{I}[P(\textbf{y},\Phi|x_i,L_0)]; \forall x_i \in U_0$
We implement DBAL as described in \cite{DBAL_gal2017deep} where probability terms in information gain is evaluated using previous equation. 


\subsection{Coreset}
Sener et al.\cite{coreset_sener2018active} exploit the geometry of data points and choose samples that provide a cover to all data points. Essentially, their algorithm tries to find a set of points (cover-points), such that distance of any data point from its nearest cover-point is minimized. They proposed two sub-optimal but efficient solutions to this NP-Hard problem: coreset-greedy and coreset-MIP (Mixed Integer programming), coreset-greedy is used to initialize coreset-MIP. For our experiments, following \cite{yoo2019learning_loss_for_AL}, we implement coreset-greedy since it achieves comparable performance while being significantly compute efficient. 

	

\subsection{Variational Adversarial Active Learning}
Sinha et al.\cite{VAAL_sinha2019variational} combined a VAE \cite{VAE_kingma} and a discriminator \cite{Goodfellow:2014:GAN:2969033.2969125} to learn a metric for AL sampling. VAE encoder is trained on both $L \text{ and } U$, and the discriminator is trained on the latent space representations of $L \text{ and } U$ to distinguish between seen ($L$) and unseen ($U$) images. Sampling function selects samples from $U$ with lowest discriminator confidence (to be seen) as measured by output of discriminator's softmax. Effectively, samples that are most likely to be unseen based on the discriminator's output are chosen. Since VAAL does not account for the end task, recent methods such as SRAAL\cite{SRAAL_zhang2020state}, TAVAAL\cite{tavaal_kim2021task} have incorporated the task awareness too.

\subsection{Ensemble Variance Ratio Learning}
Proposed by \cite{Ensembles_Beluch2018ThePO}, this is a query-by-committee (QBC) method that uses a variance ratio computed as \\
$v = 1-f_m/N$. This variance ratio select the sample set with the largest dispersion ($v$), where $N$ is the number of committee members (CNNs), and $f_m$ is the number of predictions in the modal class category. Variance ratio lies in 0--1 range and can be treated as an uncertainty measure. We note that it is possible to formulate several AL strategies using the ensemble e.g. BALD, max-entropy, etc. Variance ratio was chosen for this study because it was shown by authors to lead to superior results. For training the CNN ensembles, we train 5 models with VGG16 architecture but a different random initialization. Further, following \cite{Ensembles_Beluch2018ThePO}, the ensembles are used only for sample set selection, a separate task classifier is trained in fully-supervised manner to do image classification.

\begin{figure*}[t]
\begin{center}
    \scalebox{0.99}{
  \includegraphics[width=0.99\linewidth]{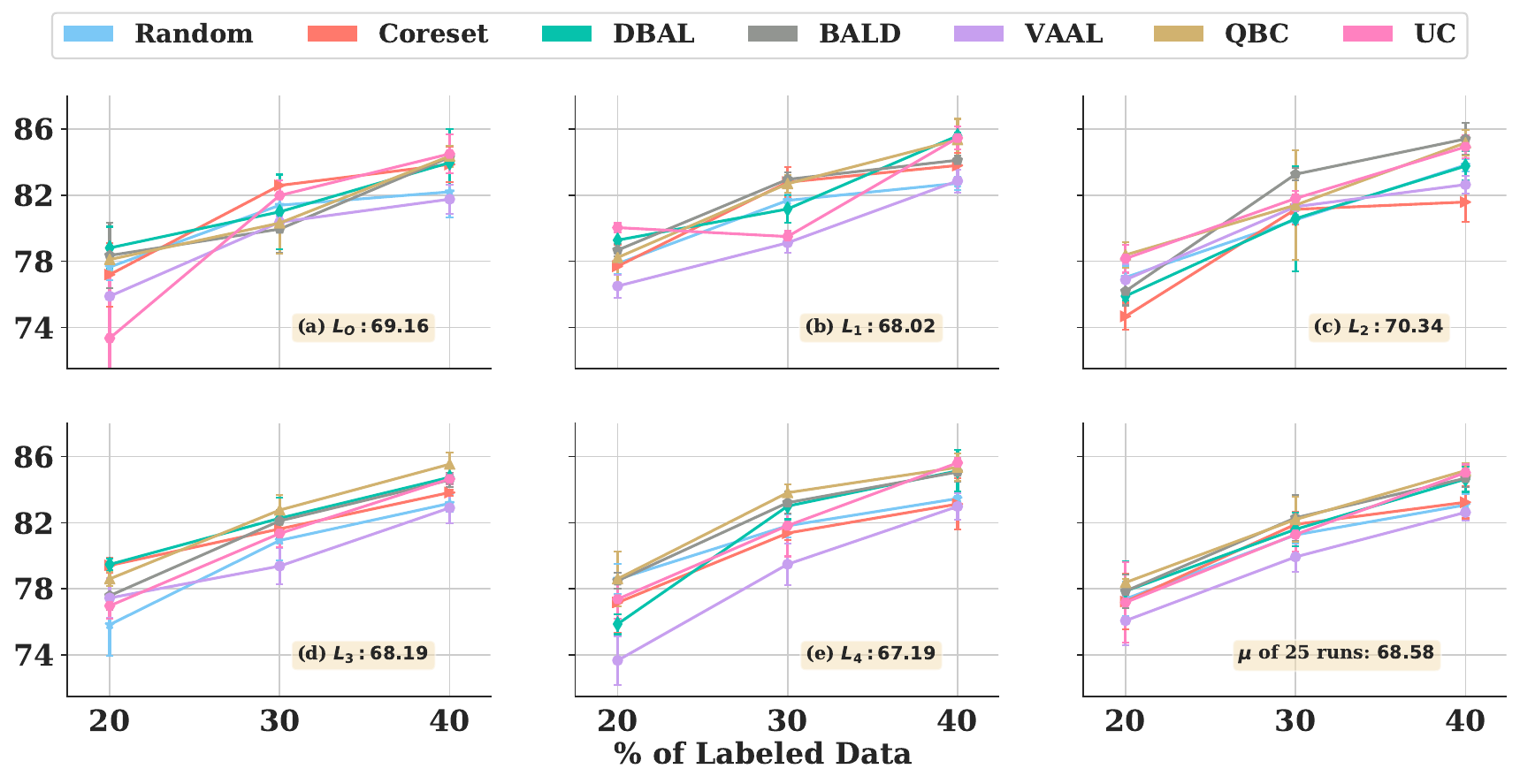}}
  \vspace{-2em}
\end{center}
    \caption{Mean accuracies achieved by AL methods compared on CIFAR10 dataset for different initial labeled sets $L_{0}, L_{1}, \cdots, L_{4}$. The mean accuracy for the base model (at 10\% labeled data) is noted inside each subplot. The model is trained $5$ times for different random initialization seeds where for the first seed we use AutoML to tune hyper-parameters and re-use these hyper-parameters for the other 4 seeds. The mean of $25$ runs (bottom right) suggest that no AL method performs consistently better than others.}
    
\label{fig:cifar_five_lSets_stats_cif10}
  \vspace{-1em}
\end{figure*}

\section{Regularization and Active Learning}\label{sec:regularization_section}
In a ML training pipeline comprising data--model--metric and training tricks, regularization can be introduced in several forms. In neural networks, regularization is commonly applied using parameter norm penalty (metric), dropout (model), or using standard data augmentation techniques such as horizontal flips and random crops (data). However, parameter norm penalty coefficients are not easy to tune and dropout effectively reduces model capacity to reduce the extent of over-fitting on the training data, and requires the drop probability to be tuned. On the other hand, several recent studies have shown promising new ways of regularizing neural networks to achieve impressive gains. While it isn't surprising that these regularization techniques help reduce generalization error, most AL studies have overlooked them. We believe this is because of a reasonable assumption that if an AL method works better than random sampling, then its relative advantage should be maintained when newer regularization techniques and training tricks are used. Since regularization is critical for low-data training regime of AL where the massively-overparameterized model can easily overfit to the limited training data, we investigate the validity of such assumptions by applying regularization techniques to the entire data--model--metric chain of neural network training.

Specifically, we employ parameter norm penalty, random augmentation (RA) \cite{cubuk2019randaugment}, stochastic weighted averaging (SWA) \cite{SWA_izmailov2018averaging}, and shake-shake (SS) \cite{shake-shake_gastaldi2017shake}. In RA, a sequence of $n$ randomly chosen image transforms are sequentially applied to the training data, with a randomly chosen distortion magnitude ($m$) which picks a value between two extremes. For details of extreme values used for each augmentation choice, we refer the reader to work of \cite{cubuk2018autoaugment}. SWA is applied on the model by first saving $e$ snapshots of model during the time-course of optimization, and then averaging the snapshots as a post-processing step. The mode of action of the regularization techniques used affect different components of the neural network training pipeline: RA is applied to data, SWA and SS is applied to model, parameter norm penalty affects the metric. In our experiments, the models which are trained using such additional regularization are referred to as strongly-regularized models (SR-models). The hyper-parameters associated with these regularization techniques as well as experiments and their results when applied to neural network training with AL-selected sample sets are discussed in \cref{Regularization}.

\section{Tuning Hyper-parameters}

The performance of deep neural networks is sensitive to the choice of hyper-parameters (e.g. learning rate, optimizer, weight decay, etc.) and there is no deterministic approach to find a combination that yields best results. Most AL methods perform grid search to find a set of hyper-parameters over the initial labeled set, and these hyper-parameters are fixed for the AL iterations \cite{coreset_sener2018active,VAAL_sinha2019variational}.
Fixing the hyper-parameters in AL iterations is questionable - with an increase in AL iterations, the size of training data increases, and the distribution changes since AL heuristics are used to draw a new set to be labeled by the oracle. Therefore, the hyper-parameters found to work well in one AL iteration may not work well at further AL iterations. To address this concern, we use AutoML at \textit{each} AL iteration in our implementation, which does 50 trials of random search over the hyper-parameters. To illustrate this point further, in 4 AL iterations for any given AL method, where initial data of 10\% is increased to 40\% with a budget size of 10\%, we train a total of 200 models and choose the best 4 (1 for each AL iteration) to report the performance. This process is repeated for each labeled set 
partition: $L_0^0$,$L_1^0$,$L_2^0$,$L_3^0$,$L_4^0$. To report the variance in accuracy at an AL iteration a labeled partition, say $L_0$,  we re-use the best hyper-parameters founded using $L_0^0$ and run on $L_0^i$, where i $\in \{ 1,2,3,4 \}$. Further details regarding the list of hyper-parameters and their range of choices is shared in the supplementary section.



\section{Implementation Details}

We perform experiments on most commonly used datasets in active learning: CIFAR10, CIFAR100, with limited additional results reported on ImageNet. 
For details on our training schedule we refer readers to the supplementary.
Given a dataset $\mathcal{D}$, we split it into train ($T_r$), validation ($V$), and test ($T_s$) sets. The train set is further divided into the initial labeled ($L_0$) and unlabeled ($U_0$) sets. A base classifier $\mathcal{B}$ is first trained, followed by iterations of sample-annotate-train process using various AL methods. Model selection is done by choosing the best performing model on the validation set. For a fair comparison, a consistent set of experimental settings is used across all methods. 
Hyper-parameters like learning rate ($lr$) and weight decay ($wd$) were tuned using AutoML with random search over 50 trials. We make use of the Optuna library \cite{optuna_akiba2019optuna} to facilitate these experiments. For ImageNet experiments, we relied on previously published hyper-parameters found using AutoML \cite{S4L_DBLP:journals/corr/abs-1905-03670}.
Models were trained from random initialization in each AL iteration for experiments that used CIFAR 10 and CIFAR 100 datasets. In case of ImageNet, the initial model trained using random drawn seed batch was trained using random initialization, and subsequent models in AL iterations were initialized using final weights from previous iteration.

\noindent \textbf{ImageNet Training Details: }We use Resnext-50 \cite{resnext_50_xie2017aggregated} as our classifier and follow the settings of \cite{S4L_DBLP:journals/corr/abs-1905-03670} \ie
Optimizer=SGD,  
$wd=3\times10^{-4}$. We train the base classifier on $L_{0}$ for $200$ epochs (300 epochs when we include SWA and RA to the training pipeline) where $lr=0.1$ with a linear warm-up schedule (for first $5$ epochs) followed by decaying the $lr$ by a factor of $10$ on epoch number: $\{140, 160, 180\}$. For AL iterations we fine-tune the best model (picked by validation set accuracy) from previous iteration for $100$ epochs where $lr=0.01$ which gets decayed by a factor of $10$ on epoch number: $\{35, 55, 80\}$. Further, we choose the best model based on a realistically small validation set (\ie $12811$ images) following \cite{S4L_DBLP:journals/corr/abs-1905-03670}. The input is pre-processed using random crops resized to $224$ x $224$ followed by horizontal flip (probability=0.5) and normalized to zero mean and one standard deviation using statistics of initial randomly drawn labeled set partition.

\noindent \textbf{Architecture \& Hyper-parameters: } All experiments are performed using the VGG16 architecture \cite{vgg16_simonyan2014very} with batchnorm \cite{batchnorm_ioffe2015batch}, unless otherwise stated. For transferability experiment (refer \cref{subsection:transfer_settings}), we use two target architectures \ie 18-layer ResNet \cite{resnet_he2016deep}, and 28-layer 2-head Wide-ResNet (WRN-28-2) \cite{wide_resnet_zagoruyko2016wide} in our experiments. Both target architectures are taken from publicly available github repository  \cite{rn18_github}, \cite{wrn_Github}.
For CIFAR10/100 models we set the number of neurons in penultimate fully-connected layer of VGG16 to $512$ as in \cite{rn18_github}. RA parameteres: N$=$the number of transformations and M$=$index of the magnitude, are tuned using AutoML. We empirically select the SWA hyperparameters as: CIFAR 10/100: SWA LR: $5\times10^{-4}$ and frequency: $50$. Imagenet: SWA LR: $1\times10^{-5}$ and frequency: $50$. 

        
    
        
        
    
    
                

\begin{figure*}
\begin{center}
    \scalebox{0.99}{
  \includegraphics[width=0.99\linewidth]{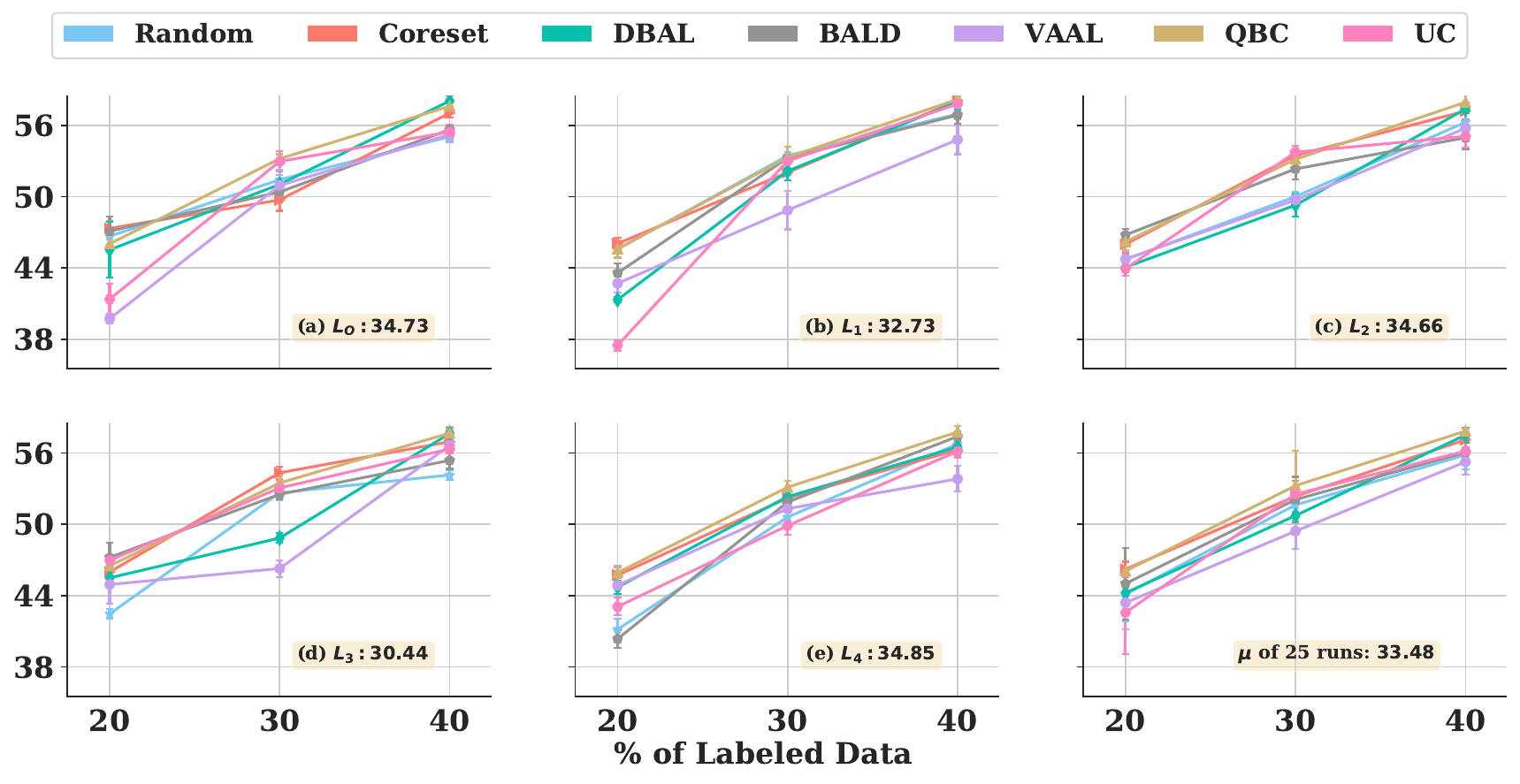}}
  \vspace{-2em}
\end{center}
    \caption{Mean accuracies achieved by AL methods compared on CIFAR100 dataset for different initial labeled sets $L_{0}, L_{1}, \cdots, L_{4}$. The mean accuracy for the base model (at 10\% labeled data) is noted inside each subplot. The model is trained $5$ times for different random initialization seeds where for the first seed we use AutoML to tune hyper-parameters and re-use these hyper-parameters for the other 4 seeds. The mean of $25$ runs (bottom right) suggest that no AL method performs consistently better than others.}
    
\label{fig:cifar_five_lSets_stats_cif100}
  \vspace{-1em}
\end{figure*}

\noindent \textbf{Implementation of AL methods:} We developed a PyTorch-based toolkit to evaluate the AL methods in a unified implementation. AL methods can be cast into two categories based on whether or not AL sampling relies on the task model (classifier network). For example, coreset uses the latent space representations learnt by task model to select the sample set, whereas VAAL relies on a separate VAE-discriminator network to select the samples, independent of the task model. In our implementation, we abstract these two approaches in a sampling function that may use the task model if required by the AL method. Each AL method was implemented using a separate sampling function, by referencing author-provided code if it was available. Using command line arguments, the toolkit allows the user to configure various aspects of training such as architecture used for task model, AL method, size of initial labeled set, size of acquisition batch, number of AL iterations, hyper-parameters for task model training and AL sampling and number of repetitions. 
\vspace{-0.39em}

\noindent \textbf{Compute:} All experiments were performed using 2 available nVidia DGX-1 servers, with each experiment utilizing 1--4 GPUs out of available 8 GPUs on each server. All codes were written in Python using PyTorch and other libraries in addition to third-party code-bases and are made available as part of the supplementary material. Models were trained over a period of many months, AutoML considerably increased time for completing each experiment.

\begin{figure*}[t]
\begin{center}
    \scalebox{0.99}{
  \includegraphics[width=0.99\linewidth]{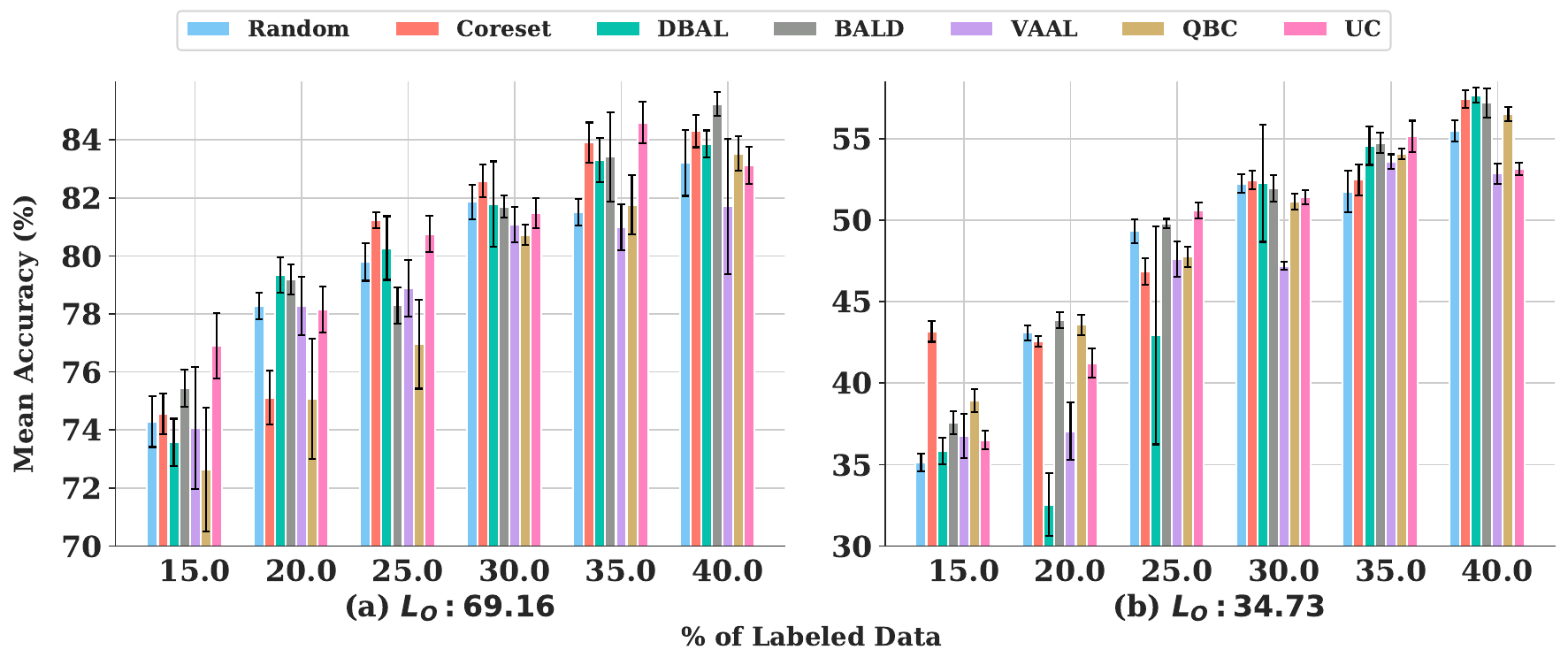}}
  \vspace{-1em}
\end{center}
    \caption{Results using $5\%$ of training data is annotated at each iteration of AL on \textbf{(a)} CIFAR10 and \textbf{(b)} CIFAR100. Mean accuracy for the base model (at 10\% labeled data) is noted at the bottom of each plot.}
    
\label{fig:cifar_budgetsize_exp}
  \vspace{-1em}
\end{figure*}

\section{Experiments and Results}

\subsection{Variance in Evaluation Metrics}
Training a neural network involves many stochastic components including parameter initialization, data augmentation, mini-batch selection, and batchnorm whose parameters change with mini-batch statistics. These elements can lead to a different optima thus resulting in varying performances across different runs of the same experiment. 
To evaluate the variance in classification accuracy caused by different initial labeled data, we draw five random initial labeled sets ($L_{0}\dots L_{4}$) with replacement. Each of these five sets were used to train the base model, initialized with random weights, 5 times; a total of 25 models were trained for each AL method to characterize variance within-sample-sets and between-sample-sets. From the results summarized in \cref{fig:cifar_five_lSets_stats_cif10} and \cref{fig:cifar_five_lSets_stats_cif100}, we make the following observations: (i) A standard deviation of \textbf{1-2}\% in accuracy among different AL methods, indicating that out of chance, it is possible to achieve seemingly better results, (ii) In contrast to previous studies, our extensive experiments indicate that no AL method performs consistently better, and random sampling baseline performs competitively. As stated previously, we believe that automatic hyper-parameter tuning repeated for each AL method at every AL iteration, while adding computation burden significantly, shows that there is marginal or no improvement achieved using AL methods compared to the random sampling baseline, (iii) Our results averaged over 25 runs in \cref{fig:cifar_five_lSets_stats_cif10} and \cref{fig:cifar_five_lSets_stats_cif100} (bottom right) indicate that no AL method performs consistently best.

\begin{table}[ht]

    \centering
    \scalebox{0.6}{
    \begin{tabular}{c|c|c|c|c}
            \toprule
            Experiments & \textgreater RSB & \textless RSB & Undetermined & \shortstack{Kruskal-Wallis\\P-Value}
                 \\
            \midrule
                C10 (20\%) & - & - & \shortstack{Coreset, DBAL, BALD,\\ VAAL, QBC, UC} & 
                $9.35\times10^{-4}$\\
            \midrule
                C10 (30\%) & - & VAAL & \shortstack{Coreset, DBAL, BALD, \\ QBC, UC} & 
                $2.67\times10^{-9}$\\
                
            \midrule
                C10 (40\%) & \shortstack{BALD, DBAL,\\ QBC} & UC & VAAL, Coreset & $8.54\times10^{-19}$\\
            
            \midrule
                C100 (20\%) & Coreset, QBC & - & \shortstack{DBAL, BALD,\\ UC} & $1.13\times10^{-8}$\\
                
            \midrule
                C100 (30\%) & QBC & VAAL & \shortstack{Coreset, DBAL,\\ BALD, UC} & $1.84\times10^{-12}$ \\
                
            \midrule
                C100 (40\%) & \shortstack{Coreset, \\ DBAL, QBC} & - & \shortstack{BALD, VAAL,\\ UC} & $5.63\times10^{-16}$ \\
            
            \bottomrule
            
            \end{tabular}
                        }
               
              \caption{Statistical Analysis of variance; C10/100 refers to CIFAR 10/100}
        \label{tab:stats_table}
   
    \end{table}

\subsection{Statistical Analysis of Variance}

In order to statistically compare the results achieved by the AL methods, we assessed the normality and variance to validate assumptions for parametric tests first. The normality assumption was tested using Kolmogorov-Smirnov test, and the assumption of homoscedasticity was tested using Levene's test. We found that normality could not be assumed for any of the six experiments in \cref{tab:stats_table}.
Using Levene's test, we found that the null hypothesis of equality of variance was rejected at $\alpha=0.05$ in 4 out of 6 experiments in \cref{tab:stats_table}. Therefore, for consistency, we used Kruskal–Wallis one-way ANOVA for all 6 experiments, and found that at least one method stochastically dominated one other method in all 6 experiments. Next, we used Games-Howell test for post-hoc multiple pairwise comparison to assess which pairs of methods differed significantly in their results. At a threshold of $\alpha=0.05$, we observed that no AL method consistently outperforms random baseline. The null hypothesis of equal mean accuracy could not be rejected for most pairwise comparisons. We did note that among the methods evaluated, QBC was superior to RSB in 4 out of 6 experiments, with other methods performing as follows: Coreset 2/6, DBAL 2/6, BALD 1/6. Our analysis strongly suggests that variance in performance metric needs to be assessed to fairly compare an AL method against RSB.

\begin{table*}[t]
\centering
\scalebox{0.8}{
\begin{tabular}{l|c c c|c c c|c c c}
\toprule
        & \multicolumn{3}{c|}{2\%}  & \multicolumn{3}{|c|}{5\%} & \multicolumn{3}{|c}{10\%} \\
\midrule
Methods & 20\%              & 30\%              & 40\%              & 20\%              & 30\%              & 40\%              & 20\%           & 30\%             & 40\% \\

\midrule
RSB  & 47.52 $\pm$ 0.58  & 52.83 $\pm$ 0.86  & 54.58 $\pm$ 0.56  &  43.43 $\pm$ 0.61 & 52.05 $\pm$ 0.6 & 57.24 $\pm$ 0.28    & 46.67 $\pm$ 0.3 & 51.43 $\pm$ 0.81   & 55.06 $\pm$ 0.35 \\ 

Coreset & 43.69 $\pm$ 0.53  & \textbf{53.37 $\pm$ 0.87}  & \textbf{57.41 $\pm$ 0.91}  &  44.31 $\pm$ 0.26   & \textbf{54.01 $\pm$ 0.87} & 56.98 $\pm$ 0.86 & \textbf{47.33 $\pm$ 0.64} & 49.73 $\pm$ 0.92   & 57.05 $\pm$ 0.4 \\ 

DBAL  & 40.32 $\pm$ 0.45  & 50.63 $\pm$ 0.55  & 57.12 $\pm$ 0.45  &  43.48 $\pm$ 1.00   & 51.45 $\pm$ 0.6    & 57.36 $\pm$ 0.73    & 45.53 $\pm$ 2.33 & 51.04 $\pm$ 0.49   & \textbf{58.06 $\pm$ 0.51} \\ 

BALD & 46.61 $\pm$ 0.37  & 51.14 $\pm$ 0.60  & 56.11 $\pm$ 0.35  &  44.47 $\pm$ 0.74   & 51.35 $\pm$ 0.61    & \textbf{58.18 $\pm$ 0.14}    & 47.1 $\pm$ 1.24 & 50.4 $\pm$ 0.88 & 55.65 $\pm$ 0.34 \\ 

VAAL  & 42.57 $\pm$ 0.89  & 49.94 $\pm$ 1.24  & 54.28 $\pm$ 0.77  &  \textbf{46.12 $\pm$ 0.96}   & 51.27 $\pm$ 0.81    & 54.41 $\pm$ 0.59    & 39.73 $\pm$ 0.43 & 50.95 $\pm$ 0.88   & 55.23 $\pm$ 0.63 \\ 

QBC & 44.86 $\pm$ 0.57  & 51.81 $\pm$ 0.53  & 56.9 $\pm$ 0.639  &  44.64 $\pm$ 0.77   & 52.16 $\pm$ 0.38    & 57.02 $\pm$ 0.21    & 46.04 $\pm$ 0.57 & \textbf{53.2 $\pm$ 0.38}   & 57.63 $\pm$ 0.49 \\ 

UC  & 44.65 $\pm$ 1.14  & 51.79 $\pm$ 0.29  & 55.81 $\pm$ 0.49  &  43.76 $\pm$ 0.19  & 52.52 $\pm$ 0.93    & 57.66 $\pm$ 0.24    & 41.37 $\pm$ 1.29 & 52.97 $\pm$ 0.83   & 55.45 $\pm$ 0.62 \\ 

\bottomrule
\end{tabular}}

\caption{Test set performance for model selected with different validation set sizes on CIFAR100. Results are average of 5 runs.}
\label{tab:valSet_exps_cifar100}
\vspace{-1em}
\end{table*}

\subsection{Differing Experimental Conditions}

Next, we compare AL methods and RSB by modifying different experimental conditions for annotation batch size, size of validation set, and class imbalance.

\noindent \textbf{Annotation Batch Size ($b$)}: Following previous studies, we experiment with annotation batch size ($b$) equal to 5\%, and 10\% of the overall sample count ($L+U$). Results in \cref{fig:cifar_budgetsize_exp} (corresponding Tab 6 in suppl.) show that our observation still holds i.e no AL method is consistently better. For example, on CIFAR10 at $40\%$ labeled data and $b=10\%$ (refer \cref{fig:cifar_five_lSets_stats_cif10} and Tab 7. in suppl.), UC outperforms all other methods but when compared to $40\%$ labeled data and $b=5\%$ (refer \cref{fig:cifar_budgetsize_exp} and Tab 6. in suppl.), BALD performs the best. Similarly on CIFAR100 at $30\%$ labeled data and $b=10\%$, RSB outperforms coreset but when compared to $30\%$ labeled data and $b=10\%$, coreset performs better. We therefore conclude that no AL method offers consistent advantage over others under different budget size settings.

\noindent \textbf{Validation Set Size}: During training, we select the best performing model on the validation set ($V$) to report the test set ($T_s$) results. To evaluate if size of $V$ can affect the conclusions drawn from comparative AL experiments, we perform experiments on CIFAR100 with three different $V$ sizes: $2\%$, $5\%$, and $10\%$ of the total samples ($L+U$).
From results in \cref{tab:valSet_exps_cifar100}, we did not observe discernible trends in accuracy with respect to the size of $V$. For instance, RSB achieves a mean accuracy of $47.5\%$, $43.4\%$, and $46.7\%$, respectively, for the best model selected using $2\%$, $5\%$ and $10\%$ of the training data as $V$. However, when the labeled data increases the range of accuracy values is not as large, indicating that training tends to suffer from less variance when higher volume of labeled data was available. For example, at $40\%$ labeled data the RSB achieves a mean accuracy of $54.58\%$, $57.24\%$, and $55.06\%$.  From these experiments, it appears that the initial AL iterations were more sensitive to size of $V$ compared to the later iterations. Furthermore, in line with previous experiment, no AL method was consistently better across AL iterations as size of $V$ changes.\\
\noindent \textbf{Class Imbalance}: Here, we evaluate the robustness of different AL methods on imbalanced data. For this, we construct $L_{0}$ on CIFAR100 dataset, to simulate long tailed distribution of classes by following a power law, where the number of samples of 100 classes are given by $\text{samples}[i] = a + b*\exp({\alpha x})$ where $i\in\{1\dots 100\}; a = 100, x = i+0.5,  \alpha = -0.046\text{ and } b = 400$. The resulting image count per class is normalized to construct a sampling distribution. Models were trained using previously described settings, with the exception of loss function which was set to weighted cross entropy. 
\begin{figure}
\begin{center}
    \scalebox{0.99}{
  \includegraphics[width=0.99\linewidth]{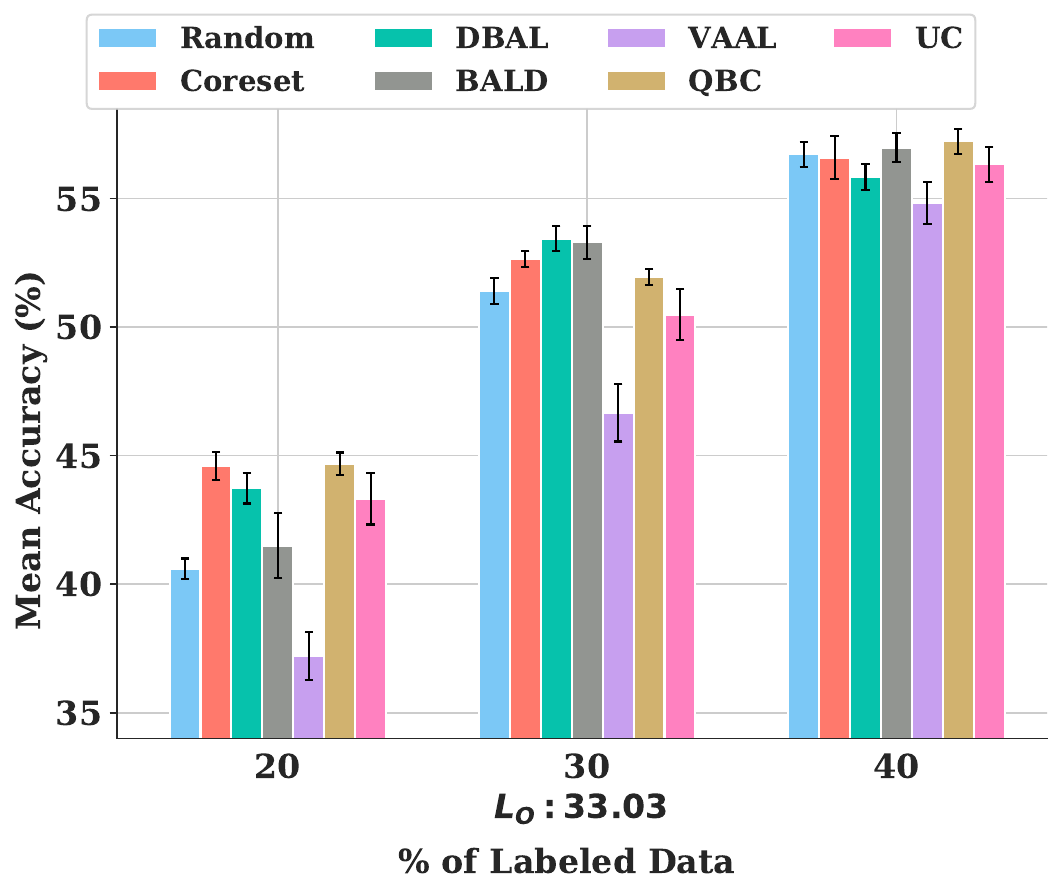}}
  \vspace{-2em}
\end{center}
    \caption{Results are average of 5 runs on imbalanced CIFAR100. The mean accuracy for the base model (at 10\% labeled data) is noted at the bottom of plot.}
    
\label{fig:class_imbalance_cif100_exp}
  \vspace{-1em}
\end{figure}

The results in \cref{fig:class_imbalance_cif100_exp} show that the difference between RSB and the best AL method reduces with more and more labeled data. More importantly, we notice that AL methods demonstrate different degree of change in the imbalanced class setting, without revealing a clear trend in the plot. From clear trend, we mean that a robust AL method is expected to perform consistently best across all fractions of the data, which is not the case in \cref{fig:class_imbalance_cif100_exp}.

\subsection{Regularization} \label{Regularization}
With the motivation stated in \cref{sec:regularization_section}, we evaluate the effectiveness of advanced regularization techniques (RA and SWA) in the context of AL using CIFAR10 and CIFAR100 datasets. We refer the models trained using such advanced regularization techniques as \textit{strongly-regularized} models (SR models) in further experiments.
We empirically observed that unlike $\ell_2-$regularization, which requires careful tuning, results using RA \& SWA were fairly robust to changes in their hyper-parameters.

\begin{figure*}
\begin{center}
    \scalebox{0.99}{
  \includegraphics[width=0.99\linewidth]{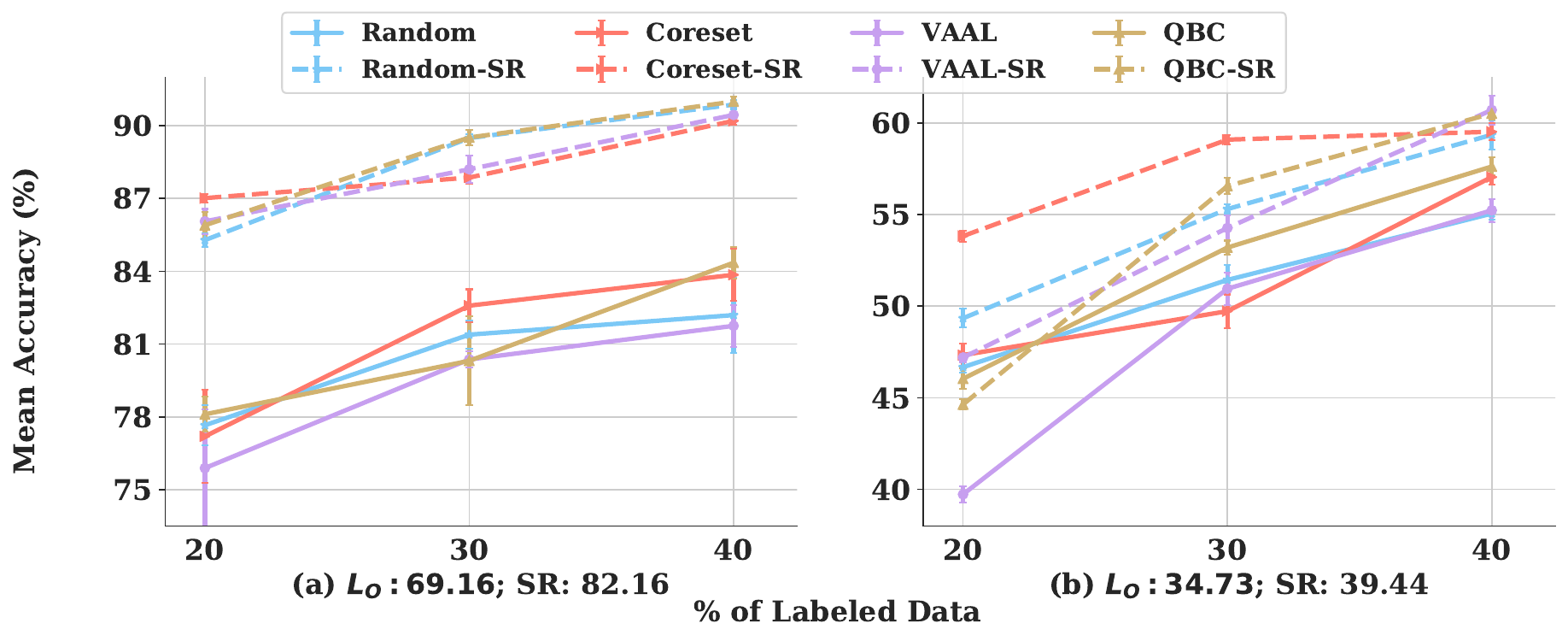}}
  \vspace{-2em}
\end{center}
    \caption{Effect of strong regularization (RA, SWA) on the test accuracy of CIFAR10(a) and CIFAR100(b). The mean accuracy for the base model (at 10\% labeled data) is noted at the bottom of each plot.}
    
\label{fig:cifar_randaug_swa_exp}
  \vspace{-1em}
\end{figure*}

\cref{fig:cifar_randaug_swa_exp} compares different AL methods with RSB on CIFAR10/100 datasets. We observe that strongly-regularized models consistently achieve significant performance gains across all AL iterations and exhibit appreciably-smaller variance across multiple runs of the experiments. Our strongly-regularized random-sampling baselines on $40\%$ labeled data achieves mean accuracy of $90.87\%$ and $59.36\%$ respectively on CIFAR10 and CIFAR100. We note that for CIFAR10, the RSB-SR model with 20\% of training data achieves $3\%$ higher accuracy compared to RSB model trained using $40\%$ of the training data. Similarly for CIFAR100, the RSB-SR $30\%$-model performs comparably to the $40\%$-RSB model. Therefore, we consider techniques under strong regularization to be a valuable addition to the low-data training regime of AL, especially given that it significantly reduces the variance in evaluation metric which can help avoid misleading conclusions.

\begin{table}[h]

    \centering
    \scalebox{0.9}{
    
    \begin{tabular}{l|c|c}
                        \toprule
                        Methods & CIFAR10 & CIFAR100 \\
                        \midrule
                        
                        RSB & 69.16 & 34.73\\
                        
                        + SWA & 74.6 & 38.06 \\
                        
                        + RA & 76.36 & 38.01 \\
                        
                        + Shake-Shake(SS) & 73.93 & 39.23 \\
                        
                        + SWA + RA & 82.16 & 39.44 \\
                        
                        + SS + SWA + RA & \textbf{84.45} & \textbf{48.92} \\
                        \bottomrule
                        \end{tabular}
                        }
               
              \caption{Individual contributions of different regularization techniques. Results correspond to the best trial (total trials=$50$) found by AutoML using 10\% of training data. Above experiments use the VGG16 except for Shake-Shake as it is restricted to the family of resnext.} 
              
            \label{tab:SWA_RANDAUG_10_percent_data}
    
    \end{table}
    \vspace{-1em}
    


  


    
  
    


\begin{figure*}
\begin{center}
    \scalebox{0.99}{
  \includegraphics[width=0.7\linewidth]{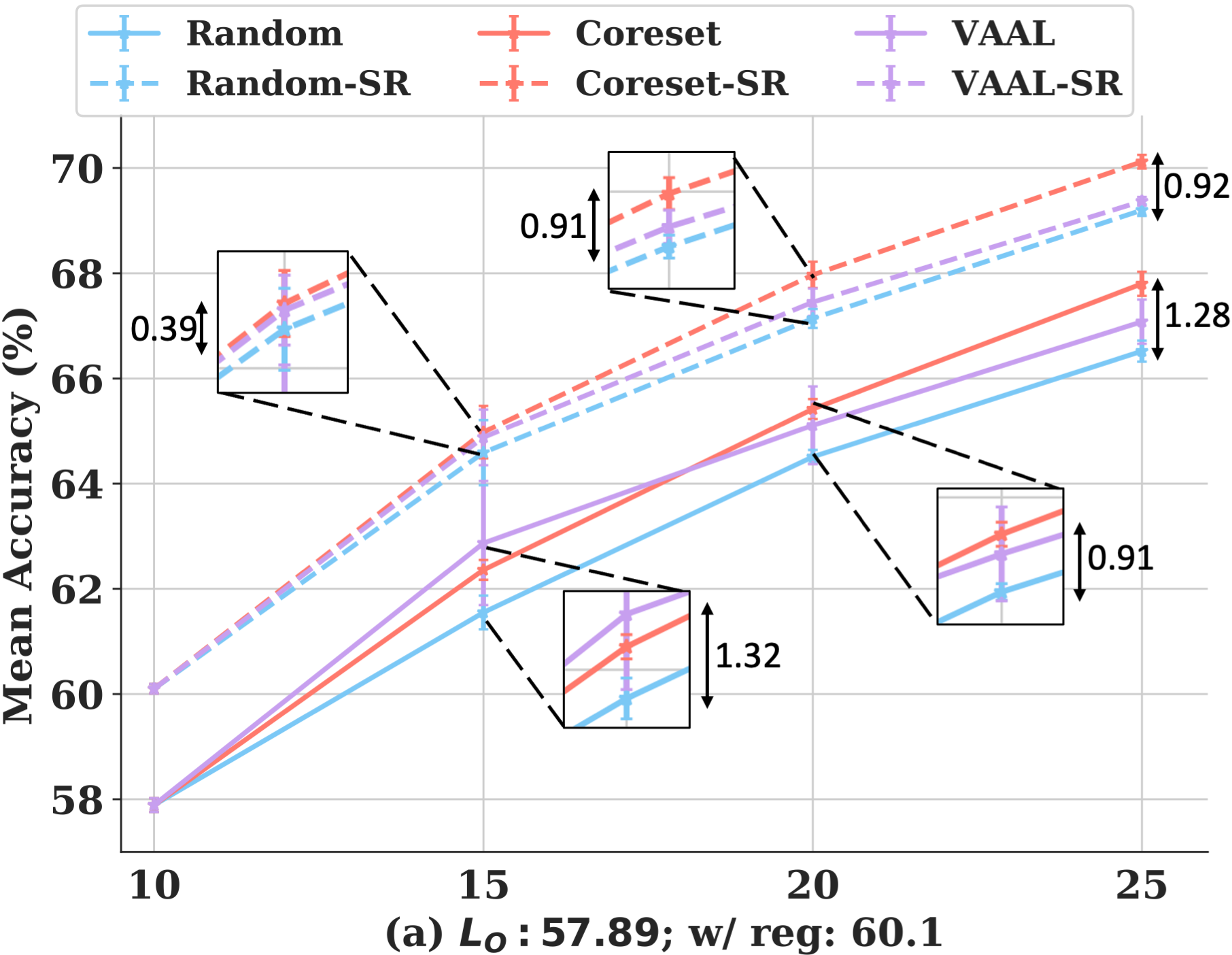}}
  \vspace{-2em}
\end{center}
    \caption{Effect of strong regularization (RA, SWA) (shown in dashed lines) on Imagenet where annotation budget is $5\%$ of training data. Reported results are averaged over 3 runs. For exact accuracies we refer readers to the Tab 5. in suppl.}
    \label{fig:imagenet_regularize}
  \vspace{-1em}
\end{figure*}

\noindent An ablative study to show individual contribution of each regularization technique towards overall performance gain is given in \cref{tab:SWA_RANDAUG_10_percent_data}. Results indicate that both RA \& SWA show a significant combined gain of $\approx12\%$ and $\approx5\%$ on CIFAR10 and CIFAR100 respectively. We also experimented with Shake-Shake (SS) \cite{shake-shake_gastaldi2017shake} in parallel to RA \& SWA, and observed that it significantly increases the runtime, and is not robust to model architectures. We therefore chose RA \& SWA over SS for strongly-regularized models.

\begin{table*}[ht]
    \begin{tabular}{cc}
         \begin{minipage}{.65\linewidth}
         \centering
         \scalebox{0.65}{
            \begin{tabular}{l|c c c|c c c|c c c|c c c} 
\toprule 
& \multicolumn{3}{c}{Source Architecture}  
 & \multicolumn{9}{c}{Target Architectures} \\ 

\toprule 
& \multicolumn{3}{c|}{VGG16} & \multicolumn{3}{c|}{WRN-28-2} & \multicolumn{3}{c|}{R18} & \multicolumn{3}{c}{R18-SR} \\
Methods & 20\%   & 30\%   & 40\%    & 20\%      & 30\%      & 40\%      & 20\%        & 30\%       & 40\%    & 20\%    & 30\%   & 40\%\\ 
\midrule 
RSB  & 77.34   & 80.91   & 82.05    & 81.3      & 81.98      & 85.68      & 75.73         & 79.69      & 80.69   & 88.69     &  90.44  & 91.8 \\
Coreset  & 76.18   & 82.24   & 85.11    & \textbf{82.32}      & 84.29      & 86.72      & \textbf{79.27}         & \textbf{82.13}      & \textbf{85.09}   & 88.48     &  91.72  & 92.94 \\
DBAL  & 78.08   & 82.42   & 85.17    & 80.79      & 84.54      & 87.87      & 75.05         & 80.69      & 82.95   & \textbf{89.41}     &  \textbf{92.23}  & \textbf{93.34} \\
VAAL  & 77.13   & 79.97   & 80.55    & 78.42      & 83.79      & 86.58      & 74.87         & 79.96      & 82.24   & 87.11     &  90.43  & 92.19 \\
QBC  & \textbf{78.63}   & \textbf{82.66}   & \textbf{85.24}    & 81.85      & \textbf{85.06}      & \textbf{87.94}      & 76.61         & 81.77      & 84.41   & 88.34     &  91.37  & 92.62 \\ \bottomrule 
\end{tabular} 
                }
               
              \caption{Transferability experiment on CIFAR10 dataset where source model is VGG16 and target model is Resnet18 (R18) and Wide Resnet-28-2 (WRN-28-2).  Test accuracies are reported corresponding to the best model trained on CIFAR10/100 dataset. For best model hyper-parameters we perform random search over 50 trials. Results with strong regularization is shown in the last column.}
              \label{tab:cifar10_transfer_experiments}

         \end{minipage}
         & 
         \begin{minipage}{.35\linewidth}
            \centering
            \scalebox{0.7}{
                        \begin{tabular}{l|cccc}
            \toprule
            Methods $\downarrow$ & $10\%$ & $20\%$ & $30\%$ & $40\%$ \\
            \midrule
            \multicolumn{5}{c}{Noise: $10\%$}\\
            \midrule
            RSB & $69.16$ & $72.08$ & $76.62$ & $80.88$\\
            \hline
            RSB-SR & $82.16$ & $84.96$ & $86.06$ & $89.13$\\
            \midrule
            \multicolumn{5}{c}{Noise: $20\%$}\\
            \midrule
            RSB & $69.16$ & $69.42$ & $75.89$ & $79.61$\\
            \hline
            RSB-SR & $82.16$ & $77.39$ & $85.9$ & $85.12$ 
            \\
            \bottomrule
            \end{tabular}
            }
           
            \caption{RSB accuracy with and without strong regularization on CIFAR10 with noisy oracle.}
            \label{tab:NOISY_cifar_10_swa_randaug}

         \end{minipage}
    \end{tabular}
    \vspace{-2em}
\end{table*}


\subsection{Active Learning on ImageNet}
Compared to CIFAR10/100, ImageNet is more challenging with larger sample count, 1000 classes and higher resolution images. In order to work with available compute resources, we compared coreset, VAAL and RSB on ImageNet. The details for training hyper-params are in supplementary material. Results with and without strong regularization are shown in \cref{fig:imagenet_regularize} (refer Tab 4 in suppl.) where mean accuracies are reported over 3 runs. Using ResNext-50 architecture \cite{resnext_50_xie2017aggregated} and following the settings of  \cite{S4L_DBLP:journals/corr/abs-1905-03670}, we achieve improved baseline performances compared to the previously reported results \cite{Ensembles_Beluch2018ThePO, VAAL_sinha2019variational}.
From \cref{fig:imagenet_regularize}, we observe that VAAL outperforms all other methods at $15\%$ data but both Coreset and Coreset-SR leads to the highest mean accuracy across other settings. We also note that similar to the observations on CIFAR datasets; (i) strong regularization helps improve performance across the methods and fraction of data, and (ii) it also reduces the performance gap between RSB and other AL methods. For example, RSB-SR model with $20\%$ data exceeds the performance achieved by both RSB and VAAL using $25\%$ data, for a saving of $\approx64000$ images with labels.

\label{subsection:transfer_settings}
\subsection{Transferability Settings}
In this experiment, we evaluated how does the accuracy compare when an active learning method chosen for one task architecture (e.g. VGG16), is used to train another task model (e.g. ResNet18). We conduct an experiment by storing the indices of sample set drawn in an AL iteration on the source network (VGG16), and use them to train the target network (ResNet18 and WRN-28-2). From \cref{tab:cifar10_transfer_experiments}, we observe the inconsistency in performance of AL methods. For example, 
on CIFAR10 with Resnet18, Coreset achieves consistently best performance. However, this observation does not hold true for strongly-regularized models. We also note that for ResNet18 target architecture, the RSB-SR model outperforms the best AL approach (coreset with ResNet18) in all the AL iterations, though DBAL-SR appears at the top with an accuracy of 93.34\% at 40\% labeled data. We therefore conclude that the AL methods are sensitive to the model architecture being used.

\section{Additional Experiments}

\label{noisyoracle}
\noindent \textbf{Noisy Oracle:} In this experiment, we sought to evaluate the stability of strongly regularized network to labels from a noisy oracle. We experimented with two levels of oracle noise by randomly permuting labels of 10\% and 20\% of samples in the set drawn by random sampling baseline at each iteration.
From results in \cref{tab:NOISY_cifar_10_swa_randaug}, we found that the drop in accuracy for the strongly-regularized model regularized was nearly half ($3\%$) compared to its complement model trained 
($6\%$) on both $30\%$ and $40\%$ data splits. Our findings suggest that the noisy pseudolabels generated for the unlabelled set $U$ by model $\phi$, when applied in conjunction with appropriate regularization, should help improve model's performance. Additional results using AL methods in this setting are shared in the supplementary material.\\

\noindent \textbf{Active Learning Sample Set Overlap:} For interested readers, we discuss the extent of overlap among the sample sets drawn by AL methods in supplementary. \\

\noindent \textbf{Optimizer Choices:} Different AL studies have reported different optimizer choices in their experiments. In this light, we analyze the optimizer chosen by AutoML and we analyze it on CIFAR10. The results are present in supplementary section. Contrary to the previous studies where the optimizer is fixed in advance, we found that both adam and sgd optimizer can sometimes work better than the other.

\section{Discussion}

\noindent \textbf{Under-Reported Baselines:} We note that several recent AL studies show baseline results that are lower than the ones reproduced in this study. Tab 1 in the supplementary summarizes our RSB results with comparisons to RSB reported by some of the recently published AL methods, under similar training settings. Based on this observation, we emphasize that comparison of AL methods must be done under a consistent set of experimental settings. Our observations confirm and provide a stronger evidence for a similar conclusion drawn in \cite{mittal2019parting}, and to a less related extent, \cite{oliver2018realistic_ssl}. Different from \cite{mittal2019parting} though, we demonstrate that: \textbf{(i)} Relative gains using AL method are found under a narrow combination of experimental conditions. \textbf{(ii)} Fixing the hyper-parameters at the start of AL can result into sub-optimal results. \textbf{(iii)} More distinctly, we show that performance gains (when they exist) are significantly lower for \textit{strongly-regularized} models.\\

\noindent \textbf{The Role of Regularization:} Regularization helps reduce generalization error and is particularly useful in training overparameterized neural networks with low data. We show that both RA \& SWA can achieve appreciable gain in performance at the expense of a small computational overhead. We observed that along with learning rate (in case of SGD), regularization was one of the key factors in reducing the error while being fairly robust to its hyperparameters (in case of RA and SWA). We also found that the margin of the gain observed with an AL method over RSB on CIFAR10/100 significantly minimize when the model is well-regularized. Strongly-regularized models also exhibited smaller variance in evaluation metric. With these observations, we recommend that AL methods be also tested using well-regularized model to ensure their robustness. Lastly, we note that there are multiple ways to regularize the data-model-metric pipeline, we focus on data and model side regularization using techniques such as RA and SWA, though it is likely that other combination of newer regularization techniques will lead to similar results. We do believe that with their simplicity and applicability to a wide variety of model (compared to shake-shake method), RA \& SWA can be effectively used in AL studies to reduce the variance.\\

\noindent \textbf{AL Methods Compared To Strong RSB:} In contrast to previous findings, well-regularized random baseline in our study was either at par with or marginally inferior to the state-of-the-art AL methods. We believe that previous studies that ran a comparison against the random baseline might have insufficiently regularized the models and/or did not tune the hyperparameters. We also observed (\cref{tab:cifar10_transfer_experiments}) that a change in model architecture can change the conclusions being drawn in comparing an AL method to a random baseline. This observations suggests that either transferability experiments should be conducted to assess if the active sets are indeed useful across architectures, or comparisons are repeated with additional architectures to study the robustness of an AL method to changes in architecture. Similarly we observed that the strong regularization achieves two goals: (i) it helps improve the performance, (ii) more importantly, it helps reduce the variance in results and reduces the performance gap between random baseline and other AL methods. The highly-sensitive nature of AL results using neural networks therefore necessitates a comprehensive suite of experimental tests.

\section{Conclusion and Proposed Guidelines}
Our extensive experiments suggest a strong need for a common evaluation platform that facilitates robust and reproducible development of AL methods. To this end, we recommend the following to ensure results are robust: 

\textbf{(i)} Experiments should be repeated under varying training settings such as model architecture and budget size, among others. \textbf{(ii)} Regularization techniques such as RA \& SWA should be incorporated into the training to ensure AL methods are able to demonstrate gains over a strong-regularized random baseline.\textbf{(iii)} Transferability experiments should be performed to test how useful AL-drawn sample sets are. Alternatively, experiments should be repeated using multiple architectures.\textbf{(iv)} To increase the reproducibility of AL results, experiments should ideally be performed using a common evaluation platform under consistent settings to minimize the sources of variation in the evaluation metric. \textbf{(v)} Snapshot of experimental settings should be shared, e.g. using a configuration file (.cfg, .json etc). \textbf{(vi)} Index sets for a public dataset used for partitioning the data into training, validation, test, and AL-drawn sets should be shared, along with the training scripts. 

\noindent In order to facilitate the use of these guidelines in AL experiments, we provide a python-based AL toolkit. We provide the index sets for the datasets used in this study that was used to partition the data into training, validation, and test sets. Lastly, all experiment configuration files are also shared as part of the toolkit.  \\   

\noindent \textbf{Societal impact and Limitations:} For some of our experiments, we use ImageNet data, which lacks gender and ethnic diversity \cite{yang2019fairer}. The models learned with this data could therefore have biased representations. Further, ImageNet has privacy concerns due to unblurred faces \cite{yang2021study}. A limitation of our work is that all our experiments are conducted for image classification. We leave other tasks such as detection and segmentation for future work.


{\small
\bibliographystyle{ieee_fullname}
\bibliography{arxiv}

\begin{thebibliography}{10}\itemsep=-1pt

\bibitem{optuna_akiba2019optuna}
Takuya Akiba, Shotaro Sano, Toshihiko Yanase, Takeru Ohta, and Masanori Koyama.
\newblock Optuna: A next-generation hyperparameter optimization framework.
\newblock In {\em Proceedings of the 25th ACM SIGKDD international conference
  on knowledge discovery \& data mining}, pages 2623--2631, 2019.

\bibitem{Ensembles_Beluch2018ThePO}
William~H. Beluch, Tim Genewein, Andreas N{\"u}rnberger, and Jan~M. K{\"o}hler.
\newblock The power of ensembles for active learning in image classification.
\newblock {\em 2018 IEEE/CVF Conference on Computer Vision and Pattern
  Recognition}, pages 9368--9377, 2018.

\bibitem{coregcn_caramalau2021sequential}
Razvan Caramalau, Binod Bhattarai, and Tae-Kyun Kim.
\newblock Sequential graph convolutional network for active learning.
\newblock In {\em Proceedings of the IEEE/CVF Conference on Computer Vision and
  Pattern Recognition}, pages 9583--9592, 2021.

\bibitem{cubuk2018autoaugment}
Ekin~D Cubuk, Barret Zoph, Dandelion Mane, Vijay Vasudevan, and Quoc~V Le.
\newblock Autoaugment: Learning augmentation policies from data.
\newblock {\em arXiv preprint arXiv:1805.09501}, 2018.

\bibitem{cubuk2019randaugment}
Ekin~D Cubuk, Barret Zoph, Jonathon Shlens, and Quoc~V Le.
\newblock Randaugment: Practical data augmentation with no separate search.
\newblock {\em arXiv preprint arXiv:1909.13719}, 2019.

\bibitem{Adv_AL_DBLP:journals/corr/abs-1802-09841}
Melanie Ducoffe and Fr{\'{e}}d{\'{e}}ric Precioso.
\newblock Adversarial active learning for deep networks: a margin based
  approach.
\newblock {\em CoRR}, abs/1802.09841, 2018.

\bibitem{DBAL_gal2017deep}
Yarin Gal, Riashat Islam, and Zoubin Ghahramani.
\newblock Deep bayesian active learning with image data.
\newblock In {\em Proceedings of the 34th International Conference on Machine
  Learning-Volume 70}, pages 1183--1192. JMLR. org, 2017.

\bibitem{shake-shake_gastaldi2017shake}
Xavier Gastaldi.
\newblock Shake-shake regularization.
\newblock {\em arXiv preprint arXiv:1705.07485}, 2017.

\bibitem{Goodfellow:2014:GAN:2969033.2969125}
Ian~J. Goodfellow, Jean Pouget-Abadie, Mehdi Mirza, Bing Xu, David
  Warde-Farley, Sherjil Ozair, Aaron Courville, and Yoshua Bengio.
\newblock Generative adversarial nets.
\newblock In {\em Advances in Neural Information Processing Systems}, pages
  2672--2680, 2014.

\bibitem{resnet_he2016deep}
Kaiming He, Xiangyu Zhang, Shaoqing Ren, and Jian Sun.
\newblock Deep residual learning for image recognition.
\newblock In {\em Proceedings of the IEEE conference on computer vision and
  pattern recognition}, pages 770--778, 2016.

\bibitem{tod_huang2021semi}
Siyu Huang, Tianyang Wang, Haoyi Xiong, Jun Huan, and Dejing Dou.
\newblock Semi-supervised active learning with temporal output discrepancy.
\newblock In {\em Proceedings of the IEEE/CVF International Conference on
  Computer Vision}, pages 3447--3456, 2021.

\bibitem{batchnorm_ioffe2015batch}
Sergey Ioffe and Christian Szegedy.
\newblock Batch normalization: Accelerating deep network training by reducing
  internal covariate shift.
\newblock {\em arXiv preprint arXiv:1502.03167}, 2015.

\bibitem{SWA_izmailov2018averaging}
Pavel Izmailov, Dmitrii Podoprikhin, Timur Garipov, Dmitry Vetrov, and
  Andrew~Gordon Wilson.
\newblock Averaging weights leads to wider optima and better generalization.
\newblock {\em arXiv preprint arXiv:1803.05407}, 2018.

\bibitem{tavaal_kim2021task}
Kwanyoung Kim, Dongwon Park, Kwang~In Kim, and Se~Young Chun.
\newblock Task-aware variational adversarial active learning.
\newblock In {\em Proceedings of the IEEE/CVF Conference on Computer Vision and
  Pattern Recognition}, pages 8166--8175, 2021.

\bibitem{VAE_kingma}
Diederik~P Kingma and Max Welling.
\newblock Auto-encoding variational bayes.
\newblock {\em arXiv preprint arXiv:1312.6114}, 2013.

\bibitem{BatchBald_DBLP:journals/corr/abs-1906-08158}
Andreas Kirsch, Joost van Amersfoort, and Yarin Gal.
\newblock Batchbald: Efficient and diverse batch acquisition for deep bayesian
  active learning.
\newblock {\em CoRR}, abs/1906.08158, 2019.

\bibitem{cifar10_dataset_krizhevsky2009learning}
Alex Krizhevsky and Geoffrey Hinton.
\newblock Learning multiple layers of features from tiny images.
\newblock Technical report, Citeseer, 2009.

\bibitem{rn18_github}
Kuangliu.
\newblock Resnet18.pytorch.
\newblock \url{https://github.com/kuangliu/pytorch-cifar}, 2017.

\bibitem{uncertainty_lewis1994sequential}
David~D Lewis and William~A Gale.
\newblock A sequential algorithm for training text classifiers.
\newblock In {\em SIGIR’94}, pages 3--12. Springer, 1994.

\bibitem{inspire1_DBLP:journals/corr/abs-1807-04801}
David Lowell, Zachary~C. Lipton, and Byron~C. Wallace.
\newblock How transferable are the datasets collected by active learners?
\newblock {\em CoRR}, abs/1807.04801, 2018.

\bibitem{wrn_Github}
Meliketoy.
\newblock wide-resnet.pytorch.
\newblock \url{https://github.com/meliketoy/wide-resnet.pytorch}, 2017.

\bibitem{mittal2019parting}
Sudhanshu Mittal, Maxim Tatarchenko, Özgün Çiçek, and Thomas Brox.
\newblock Parting with illusions about deep active learning, 2019.

\bibitem{oliver2018realistic_ssl}
Avital Oliver, Augustus Odena, Colin~A Raffel, Ekin~Dogus Cubuk, and Ian
  Goodfellow.
\newblock Realistic evaluation of deep semi-supervised learning algorithms.
\newblock In {\em Advances in Neural Information Processing Systems}, pages
  3235--3246, 2018.

\bibitem{inspire2_prabhu2019sampling}
Ameya Prabhu, Charles Dognin, and Maneesh Singh.
\newblock Sampling bias in deep active classification: An empirical study.
\newblock {\em arXiv preprint arXiv:1909.09389}, 2019.

\bibitem{coreset_sener2018active}
Ozan Sener and Silvio Savarese.
\newblock Active learning for convolutional neural networks: A core-set
  approach.
\newblock In {\em International Conference on Learning Representations}, 2018.

\bibitem{vgg16_simonyan2014very}
Karen Simonyan and Andrew Zisserman.
\newblock Very deep convolutional networks for large-scale image recognition.
\newblock {\em arXiv preprint arXiv:1409.1556}, 2014.

\bibitem{VAAL_sinha2019variational}
Samarth Sinha, Sayna Ebrahimi, and Trevor Darrell.
\newblock Variational adversarial active learning.
\newblock {\em arXiv preprint arXiv:1904.00370}, 2019.

\bibitem{BGALD_DBLP:journals/corr/abs-1904-11643}
Toan Tran, Thanh{-}Toan Do, Ian~D. Reid, and Gustavo Carneiro.
\newblock Bayesian generative active deep learning.
\newblock {\em CoRR}, abs/1904.11643, 2019.

\bibitem{resnext_50_xie2017aggregated}
Saining Xie, Ross Girshick, Piotr Doll{\'a}r, Zhuowen Tu, and Kaiming He.
\newblock Aggregated residual transformations for deep neural networks.
\newblock In {\em Proceedings of the IEEE conference on computer vision and
  pattern recognition}, pages 1492--1500, 2017.

\bibitem{yang2019fairer}
Kaiyu Yang, Klint Qinami, Li Fei-Fei, Jia Deng, and Olga Russakovsky.
\newblock Towards fairer datasets: Filtering and balancing the distribution of
  the people subtree in the imagenet hierarchy.
\newblock In {\em Proceedings of the 2020 conference on fairness,
  accountability, and transparency}, pages 547--558, 2020.

\bibitem{yang2021study}
Kaiyu Yang, Jacqueline Yau, Li Fei-Fei, Jia Deng, and Olga Russakovsky.
\newblock A study of face obfuscation in imagenet.
\newblock {\em arXiv preprint arXiv:2103.06191}, 2021.

\bibitem{yoo2019learning_loss_for_AL}
Donggeun Yoo and In~So Kweon.
\newblock Learning loss for active learning.
\newblock In {\em Proceedings of the IEEE Conference on Computer Vision and
  Pattern Recognition}, pages 93--102, 2019.

\bibitem{wide_resnet_zagoruyko2016wide}
Sergey Zagoruyko and Nikos Komodakis.
\newblock Wide residual networks.
\newblock {\em arXiv preprint arXiv:1605.07146}, 2016.

\bibitem{S4L_DBLP:journals/corr/abs-1905-03670}
Xiaohua Zhai, Avital Oliver, Alexander Kolesnikov, and Lucas Beyer.
\newblock S\({}^{\mbox{4}}\)l: Self-supervised semi-supervised learning.
\newblock {\em CoRR}, abs/1905.03670, 2019.

\bibitem{SRAAL_zhang2020state}
Beichen Zhang, Liang Li, Shijie Yang, Shuhui Wang, Zheng-Jun Zha, and Qingming
  Huang.
\newblock State-relabeling adversarial active learning.
\newblock In {\em Proceedings of the IEEE/CVF Conference on Computer Vision and
  Pattern Recognition}, pages 8756--8765, 2020.

\end{thebibliography}
}


\newpage
\onecolumn
\setcounter{section}{0}
\setcounter{table}{0}
\section{Supplementary Section}
In this section we mention interesting observations and the training details which were used to report the experiments in the main paper. In addition to this, we also 
discuss the training schedule and additional experiments (for example transferability experiment on CIFAR100) which we could not fit in the main paper due to space constraints.

\subsection{Underreported Baselines}

\begin{table*}[ht]
    \centering
    \scalebox{0.9}{
                \begin{tabular}{l|c|c|c|c|c}
                \toprule
                
                Methods & $10$\% & $20$\% & $30$\% & $40$\% & Model \\
                \midrule
                \multicolumn{6}{c}{CIFAR10} \\
                \midrule
                QBC & $74$ & $82.5$ & - & - & DenseNet121 \\
                \hline
                VAAL & $61.35$ & $68.17$ & $72.26$ & $75.99$ & VGG16 \\
                \hline
                Coreset & $60$ & $68$ & $71$ & $74$ & VGG16 \\
                \hline
                RSB(ours) & $69.16$ & $77.34$ & $80.91$ & $82.05$ & VGG16 \\
                \hline
                RSB-SR(ours) & $82.16$ & $85.09$ & $89.43$ & $91.16$ & VGG16 \\
                \hline
                
                LLAL
                & $81$ & $87$ & - & - & ResNet18 \\
                \hline
                CoreGCN & $80$ & $85.5$ & - & - & ResNet18 \\
                \hline
                TA-VAAL
                & $81$ & $87.5$ & - & - & ResNet18 \\
                \hline
                RSB-SR(ours) & $\mathbf{84.69}$ & $\mathbf{88.45}$ & $\mathbf{89.98}$ & $\mathbf{92.29}$ & ResNet18 \\
                \bottomrule
                \end{tabular}

                }
               
                \caption{Reported Random Baseline vs our RSB results. We denote RSB results with strong regularization by RSB-SR.} 

                \label{tab:baselines_vs_reported_results}
                \vspace{-1em}
\end{table*}

In this section we analyze our random baseline (RSB) results with the random baselines reported by published methods in AL literature. From \cref{tab:baselines_vs_reported_results}, it is evident that our strongly-regularized settings along with hyper-parameters tuned using AutoML yields strong baseline. 



        
    
        
        
    
    

\subsection{Training Algorithm}

\begin{algorithm}[ht]
    \SetAlgoLined
	\caption{AL Training Schedule}
	\label{alg:train_schedule}
	
	\begin{algorithmic}[1]
	    \STATE Input $AL_{iter}$, Budget size $k$ and Oracle, $\mathcal{A}$
        \STATE Split $\mathcal{D} \rightarrow \{T_{r}, T_{s}, V\} $
        \STATE Split $T_{r} \rightarrow \{L_{0}^{0}, U_{0}^{0}\} $
        \STATE Train a base classifier, $\mathcal{B}$ using only $L_{0}^{0}$
        \STATE $\phi = \mathcal{B}$
        \WHILE{ i $\in \{0\dots AL_{iter}\}$}
            \STATE sample $\{ x_j\}_{j=1}^{k}\in U_{0}^{i}$ using $\Psi(L_{0}^{i}, U_{0}^{i}, \phi)$
            \STATE $\{x_j,y_j\}_{j=1}^{k} \leftarrow \{x_j,\mathcal{A}(x_j)\}_{j=1}^{k} $ 
            \STATE $L_{0}^{i} \leftarrow L_{0}^{i} \cup \{x_j,y_j\}_{j=1}^{k}$
            \STATE $U_{0}^{i} \leftarrow U_{0}^{i} \setminus \{x_j,y_j\}_{j=1}^{k}$
            \STATE $\phi \leftarrow $Initialize randomly
            \WHILE{convergence}
                \STATE Train $\phi$ using only $L_{0}^{i}$
            \ENDWHILE
        \ENDWHILE
	\end{algorithmic}
\end{algorithm}

For all reported experiments in the main paper we followed the algorithm described in \cref{alg:train_schedule}

\subsection{Auto-ML Hyper-parameters}
\label{sup:AutoMLhyperparamssection}

Here we enlist our hyper-parameters tuned using AutoML. To implement AutoML we used optuna framework extensively in our codebase.
\begin{itemize}
    \item Learning rate: log-scale in range $[10^{-5}, 10^{-2})$
    \item Weight Decay : log-scale in range $[10^{-8}, 10^{-3})$
    \item Batch Size: Categorical values from [8,16,32....1024]
    \item Optimizer: Categorical values from [SGD, ADAM]
    \item Number of Transformation in randaug (RA\_N) : Categorical values from [1,2,3,....15]
    \item Magnitude of Transformation in randaug (RA\_M) : Categorical values from [1,2,3,....8]
\end{itemize}

\subsection{Transferability Experiment}
We mainly used three different architectures for classifier model \ie VGG16, ResNet18 (R18) and Wide ResNet-28-2 (WRN)\footnotemark{\value{0}}. 
\footnotetext{\smaller\relax All Model definitions in AL Toolkit has been provided as a supplementary material}
The VGG network was used as a source model whereas other two networks are used for target models. The results for CIFAR100 are reported in Table \ref{tab:cifar100_transfer_experiments} which are achieved when we replace all the relu activations with leaky relu (negative slope set to $0.2$) following (Oliver \etal, $2018$). We found CIFAR100 results to be significantly better with leaky relu activation, however, the same change does not affect the performance of CIFAR10.

\begin{table}[ht]
\begin{center}
\begin{tabular}{l|c c c|c c c|c c c}

\toprule
        & \multicolumn{3}{c}{Source Model} & \multicolumn{6}{c}{Target Model}\\
        \midrule
& \multicolumn{3}{c|}{VGG16}  & \multicolumn{3}{c|}{WRN-28-2} & \multicolumn{3}{c}{R18-SR} \\
Methods $\downarrow$  & 20\%   & 30\%   & 40\%    & 20\%      & 30\%      & 40\%      & 20\%        & 30\%       & 40\% \\
 \midrule
Random  & $46.72$   & $50.63$   & $55.27$    & $47.87$      & $56.53$      & $57.84$     & $60.17$  & $64.8$      & $69.33$  \\
Coreset  & $\textbf{48.2}$   & $49.5$   & $\textbf{56.99}$    & $\textbf{51.25}$     & $\textbf{58.39}$     & $\textbf{60.56}$      & $58.76$       & $65.40$      & $69.12$  \\
VAAL  & $39.32$   & $52.17$   & $55.73$    & $49.13$      & $57.72$      & $55.71$      & $59.76$         & $61.36$      & $67.15$   \\
QBC  & $46.53$   & $\textbf{53.16}$   & $55.54$    & $49.02$      & $53.51$      & $57.05$      & $\textbf{61.06}$         & $\textbf{66.92}$      & $\textbf{69.83}$   \\

\bottomrule
\end{tabular}
\end{center}

\caption{Transferability experiment on CIFAR100 dataset where source model is VGG16. The reported numbers are test accuracies corresponding to the best trained on CIFAR100 dataset. For best model hyper-parameters we perform random search over 50 trials(so for 4 AL iterations; we train 200 models in total). For this experiment we replace all relu activations with leaky relu (negative slope set to $0.2$). 
}
\label{tab:cifar100_transfer_experiments}
\end{table}

\subsection{Optimizer settings}

Different AL studies have reported different optimizer choices in their experiments. In this light, we analyze the optimizer chosen by AutoML and we analyze it on CIFAR10. The results are present in Table \ref{tab:Optim_choices} of supplementary section. Contrary to the previous works where the optimizer is fixed in advance, we found that both Adam and SGD can sometimes work better than the other.

\begin{table}[h]
\begin{center}
\scalebox{0.9}{
                
                \begin{tabular}{ l|c|c|c }
                
                \toprule
                Optimizers & $20$\% & $30$\% & $40$\% \\
                \midrule
                \multicolumn{4}{c}{CIFAR10}\\
                \midrule
                
                SGD & 4 & 3 & 5 \\
                \hline
                ADAM & 10 & 11 & 9 \\
                \midrule
                \multicolumn{4}{c}{CIFAR100}\\
                \midrule
                
                SGD & 10 & 10 & 9 \\
                \hline
                ADAM & 4 & 4 & 5\\
                \bottomrule
                \end{tabular}
                
                }
                
                \caption{Analyzing best optimizer chosen by AutoML during random search over 50 trials for all the AL methods (VGG16 classifier) on CIFAR 10. As we implement 7 AL methods in both standard and strongly-regularized settings; so at each AL iteration we have a total of 14 best optimizers chosen.}
                \label{tab:Optim_choices}

\end{center}
\end{table}

\subsection{Noisy Oracle Experiments}
In conjunction to RSB baselines (presented in main paper), we report performance of AL methods under noisy labels in active sets. The results are reported in \cref{tab:noisy_oracle} where we make the following observations: \textbf{(i)} it is quite evident that strongly-regularized model improves performance even in label corruptions scenarios. \textbf{(ii)} No AL method consistently outperforms the simple RSB baseline. \textbf{(iii)} Strong-regularization help reduce the performance difference between RSB and best AL method at a particular data split.

\begin{table*}[h]
\centering
\begin{tabular}{l|cccc|cccc}
\toprule
 & \multicolumn{4}{c|}{\textbf{\shortstack{without \\strong-regularization}}} & \multicolumn{4}{c}{\textbf{\shortstack{with \\strong-regularization}}}\\
\midrule
\textbf{Methods $\downarrow$} & $10\%$ & $20\%$ & $30\%$ & $40\%$ & $10\%$ & $20\%$ & $30\%$ & $40\%$ \\
\midrule
\multicolumn{9}{c}{Noise: $10\%$}\\
\midrule
RSB & 69.16 & 72.08 & 76.62 & 80.88 & 82.16 & 84.96 & 86.06 & 89.13 \\
Coreset & 69.16 & 75.97 & 80.07 & 82.78 & 82.16 & 82.99 & 88.14 & 90.31 \\
DBAL & 69.16 & \textbf{76.98} & \textbf{80.5} & \textbf{84.4} & 82.16 & 85.04 & 88.04 & \textbf{90.44} \\
BALD & 69.16 & 75.29 & 80.24 & 84.04 & 82.16 & 82.15 & \textbf{88.24} & 89.45 \\
VAAL & 69.16 & 73.85 & 77.35 & 79.82 & 82.16 & \textbf{85.32} & 86.57 & 89.53 \\
QBC & 69.16 & 75.64 & 77.87 & 80.53 & 82.16 & 85.25 & 87.39 & 88.68 \\
UC & 69.16 & 75.94 & 80.42 & 81.92 & 82.16 & 82.61 & 85.19 & 88.62 \\

\midrule
\multicolumn{9}{c}{Noise: $20\%$}\\
\midrule
RSB & 69.16 & 69.42 & 75.89 & 79.61 & 82.16 & 77.39 & 85.9 & 85.12 \\
Coreset & 69.16 & 71.13 & 76.44 & \textbf{80.07} & 82.16 & 80.05 & 88.05 & 88.32 \\
DBAL & 69.16 & 71.26 & 76.24 & 82.2 & 82.16 & 81.31 & 83.67 & 91.14 \\
BALD & 69.16 & 70.34 & \textbf{77.18} & 79.86 & 82.16 & \textbf{85.26} & \textbf{88.52} & \textbf{91.21} \\
VAAL & 69.16 & 70.13 & 74.94 & 76.42 & 82.16 & 82.39 & 82.66 & 88.31 \\
QBC & 69.16 & 71.18 & 76.52 & 77.78 & 82.16 & 83.17 & 84.68 & 85.62 \\
UC & 69.16 & \textbf{71.53} & 75.48 & 78.48 & 82.16 & 84.57 & 83.00 & 88.42 \\

\bottomrule
\end{tabular}

\caption{Mean accuracy on noisy oracle experiments on CIFAR10 with (n=3) repeated trials where the best hyper-parameters were found using the random search over 50 trials. We note that the noise is added in active sets drawn by AL methods. The strong-regularization experiments involve SWA and RA techniques.}
\label{tab:noisy_oracle}
\end{table*}

\subsection{Overlap in the active set}
\label{supp:section_overlap_active_set}
For the interested readers we plot the overlap in CIFAR10 active set sampled in the first AL iteration. As we do five runs for a labeled set partition, we therefore report the average overlap in Figure \ref{fig:overlap_cif10}.

\begin{table*}[ht]
  \begin{center}

  \begin{tabular}{cc}

    \begin{minipage}{.6\linewidth}
    \centering
    \resizebox{0.8\linewidth}{!}{
    \begin{tabular}{l|cccc}
                \toprule
                Methods $\downarrow$ & $10\%$ & $15\%$ & $20\%$ & $25\%$ \\
                \midrule
                \multicolumn{5}{c}{without RA + SWA}\\
                \midrule
                RSB & 57.89 $\pm$ 0.13 & 61.55 $\pm$ 0.32 & 64.51 $\pm$ 0.13 & 66.52 $\pm$ 0.2 \\
                
                Coreset & 57.89 $\pm$ 0.13 & 62.36 $\pm$ 0.19 & \textbf{65.42} $\pm$ \textbf{0.19} & \textbf{67.8} $\pm$ \textbf{0.23}\\
                
                VAAL & 57.89 $\pm$ 0.13 & \textbf{62.87} $\pm$ \textbf{1.18} & 65.11 $\pm$ 0.74 & 67.08 $\pm$ 0.42\\
                \midrule
                \multicolumn{5}{c}{with RA + SWA}\\
                \midrule
                
                RSB & 60.1 $\pm$ 0.09 & 64.59 $\pm$ 0.62 & 67.14 $\pm$ 0.18 & 69.2 $\pm$ 0.11 \\
                
                Coreset & 60.1 $\pm$ 0.09 & \textbf{64.98} $\pm$ \textbf{0.50} & \textbf{67.97} $\pm$ \textbf{0.25} & \textbf{70.12} $\pm$ \textbf{0.13}\\

                VAAL & 60.1 $\pm$ 0.09 & 64.88 $\pm$ 0.53 & 67.45 $\pm$ 0.26 & 69.37 $\pm$ 0.08\\

                \bottomrule
                \end{tabular}}

                \caption{Effect of RA and SWA on ImageNet where annotation budget is $5\%$ of training data. Reported results are averaged over 3 runs.}
                \label{tab:RA_SWA_Imagenet_exp}

    \end{minipage}
    
    &
    
    \begin{minipage}{.4\linewidth}
\begin{center}
    \scalebox{0.8}{
  \includegraphics[width=0.99\linewidth]{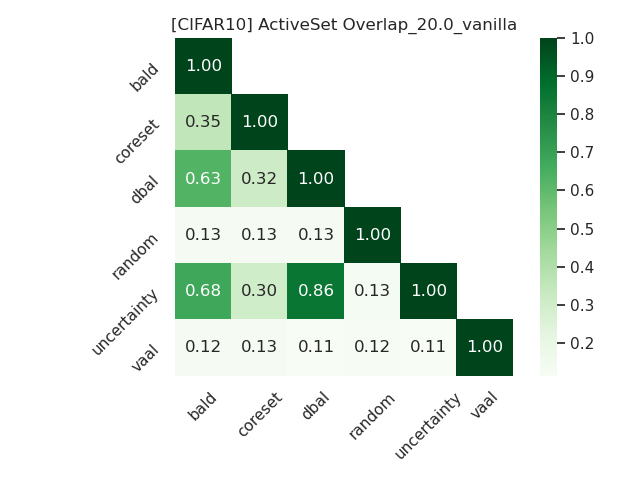}}
\end{center}
\vspace{-5mm}
  \captionof{figure}{Overlap in CIFAR10 active set which is sampled during the first AL iteration. 
  }
\label{fig:overlap_cif10}

    \end{minipage}
    
  \end{tabular}
  \end{center}
\end{table*}

\subsection{Annotation Batch Size}
Here we present the results for CIFAR10 and CIFAR100 in Table \ref{tab:CIFAR_exp_budget_5k} for the experiment where annotation batch size is $5\%$ relative to training data.

\begin{table*}[ht]

\centering
\resizebox{0.85\linewidth}{!}{ 
\begin{tabular}{l|cccccc}
\toprule
Methods $\downarrow$ & $15\%$ & $20\%$ & $25\%$ & $30\%$ & $35\%$ & $40\%$ \\
\midrule
 & \multicolumn{6}{c}{CIFAR10}\\
\midrule

RSB &  74.30 $\pm$ 0.88 & 78.27 $\pm$ 0.47 & 79.79 $\pm$ 0.64 & 81.86 $\pm$ 0.60 & 81.50 $\pm$ 0.45 & 83.21 $\pm$ 1.14\\

Coreset &  74.56 $\pm$ 0.70 & 75.11 $\pm$ 0.92 & \textbf{81.23} $\pm$ \textbf{0.27} & \textbf{82.58} $\pm$ \textbf{0.57} & 83.9 $\pm$ 0.70 & 84.30 $\pm$ 0.56\\

DBAL &  73.58 $\pm$ 0.81 & \textbf{79.33} $\pm$ \textbf{0.61} & 80.27 $\pm$ 1.10 & 81.78 $\pm$ 1.47 & 83.30 $\pm$ 0.75 & 83.86 $\pm$ 0.47\\

BALD &  \textbf{75.43} $\pm$ \textbf{0.63} & 79.19 $\pm$ 0.51 & 78.29 $\pm$ 0.63 & 81.69 $\pm$ 0.38 & 83.42 $\pm$ 1.54 & \textbf{85.23} $\pm$ \textbf{0.41}\\

VAAL &  74.07 $\pm$ 2.11 & 78.28 $\pm$ 1.00 & 78.88 $\pm$ 0.97 & 81.07 $\pm$ 0.61 & 80.98 $\pm$ 0.79 & 81.72 $\pm$ 2.33\\

QBC &  72.63 $\pm$ 2.14 & 75.07 $\pm$ 2.07 & 76.95 $\pm$ 1.52 & 80.72 $\pm$ 0.34 & 81.76 $\pm$ 1.03 & 83.53 $\pm$ 0.59\\

UC &  76.90 $\pm$ 1.12 & 78.14 $\pm$ 0.79 & 80.75 $\pm$ 0.62 & 81.47 $\pm$ 0.53 & \textbf{84.60} $\pm$ \textbf{0.71} & 83.13 $\pm$ 0.64\\

\midrule
 & \multicolumn{6}{c}{CIFAR100}\\
\midrule

RSB &  35.15 $\pm$ 0.55 & 43.10 $\pm$ 0.46 & 49.33 $\pm$ 0.73 & 52.24 $\pm$ 0.56 & 51.76 $\pm$ 1.29 & 55.49 $\pm$ 0.64\\

Coreset &  43.19 $\pm$ 0.65 & 42.58 $\pm$ 0.32 & 46.85 $\pm$ 0.83 & \textbf{52.47} $\pm$ \textbf{0.58} & 52.48 $\pm$ 0.93 & 57.45 $\pm$ 0.54\\

DBAL &  35.83 $\pm$ 0.83 & 32.54 $\pm$ 1.92 & 42.93 $\pm$ 6.69 & 52.27 $\pm$ 3.59 & 54.58 $\pm$ 1.18 & \textbf{57.68} $\pm$ \textbf{0.46}\\

BALD &  37.55 $\pm$ 0.70 & \textbf{43.86} $\pm$ \textbf{0.48} & 49.79 $\pm$ 0.29 & 51.96 $\pm$ 0.81 & 54.75 $\pm$ 0.63 & 57.20 $\pm$ 0.90\\

VAAL &  36.75 $\pm$ 1.36 & 37.05 $\pm$ 1.78 & 47.62 $\pm$ 1.07 & 47.20 $\pm$ 0.25 & 53.61 $\pm$ 0.44 & 52.87 $\pm$ 0.63\\

QBC &  \textbf{38.91} $\pm$ \textbf{0.70} & 43.57 $\pm$ 0.62 & 47.76 $\pm$ 0.61 & 51.16 $\pm$ 0.49 & 54.06 $\pm$ 0.33 & 56.51 $\pm$ 0.42\\

UC &  36.52 $\pm$ 0.55 & 41.23 $\pm$ 0.89 & \textbf{50.59} $\pm$ \textbf{0.50} & 51.42 $\pm$ 0.42 & \textbf{55.14} $\pm$ \textbf{0.97} & 53.15 $\pm$ 0.36\\
\bottomrule
\end{tabular}}
\caption{Mean Accuracy and Standard Deviation on CIFAR10/100 test set with annotation size as $5\%$ of training set. Results reported are averaged over 5 runs where hyper-parameters are tuned in the first run using AutoML random search over 50 trials.}
\label{tab:CIFAR_exp_budget_5k}
\end{table*}

\subsection{Unexplained performance degradation}

In this section we discuss an counter-intuitive observation seen during AL iterations \ie even with the increase in the labeled data, we sometimes observed the model performance (classification accuracy) degrading. More importantly, this observation was seen across different AL methods and datasets. For example on CIFAR10 from 20\% to 30\% AL cycle, the uncertainty method degrades its performance by \textbf{0.54\%} (refer \cref{tab:cif10_l0_1_supp}). Similarly on CIFAR100 and CIFAR10 from 30\% to 35\% AL cycle, the coreset and vaal method degrades its performance by \textbf{0.01\%} and \textbf{0.09\%} respectively (refer \cref{tab:CIFAR_exp_budget_5k}). Infact during our initial experiments without AutoML and strong-regularization, we observed such behaviour more frequent along-with high variance in accuracy and inconsistent ordinal ranking (by accuracy) across fractions of the data. These observations led us to employ AutoML and  strong regularization, which helped reduce variance. We hypothesize that the performance drop could occur through a suboptimal active set selection by the AL method, as we do not interfere with active sets or settings used for AutoML (best of 50 experiments).

\section{Additional Results}
In the last we present the exact accuracies which were used to plot the Figure 1 and Figure 2 in main paper. \cref{tab:CIFAR10_tab7_supp} to \cref{tab:CIFAR10_tab11_supp} reports the test accuracies for CIFAR10 dataset and \cref{tab:CIFAR100_tab12_supp} to \cref{tab:CIFAR100_tab16_supp} reports the test accuracies for CIFAR100 dataset.

\begin{table}[ht]
    \begin{center}
        \begin{tabular}{c c}
            
        \begin{minipage}{0.5\linewidth}   
        \centering
        \resizebox{0.85\linewidth}{!}{
        
        \begin{tabular}{l|cccc}
        \toprule
        Methods & $20\%$ & $30\%$ & $40\%$ \\
        \midrule
        RSB & 77.65 $\pm$ 0.82 & 81.39 $\pm$ 0.59 & 82.19 $\pm$ 1.55\\
        
        Coreset & 77.19 $\pm$ 1.93 & \textbf{82.58} $\pm$ \textbf{0.67} & 83.86 $\pm$ 1.08\\
        DBAL & \textbf{78.81} $\pm$ \textbf{1.28} & 80.99 $\pm$ 2.25 & 83.96 $\pm$ 2.01 \\
        BALD & 78.35 $\pm$ 1.98 & 79.95 $\pm$ 1.43 & 84.29 $\pm$ 0.25\\
        VAAL & 75.89 $\pm$ 2.41 & 80.37 $\pm$ 0.34 & 81.75 $\pm$ 0.87\\
        QBC & 78.10 $\pm$ 0.73 & 80.31 $\pm$ 1.83 & 84.35 $\pm$ 0.64\\
        UC & 73.35 $\pm$ 4.84 & 81.98 $\pm$ 0.93 & \textbf{84.49} $\pm$ \textbf{1.18}\\
        
        \bottomrule
        \end{tabular}
        }
  
        \caption{CIFAR10 Test Accuracy on L$_{0}^{0}$. The base model accuracy is $69.16$.}
        
        \label{tab:CIFAR10_tab7_supp}

    \end{minipage} 
    
    &
    
    \begin{minipage}{0.5\linewidth}
    \centering
    \resizebox{0.85\linewidth}{!}{
        \begin{tabular}{l|cccc}
        \toprule
        Methods & $20\%$ & $30\%$ & $40\%$ \\
        \midrule
        RSB & 77.85 $\pm$ 0.65 & 81.68 $\pm$ 0.39 & 82.71 $\pm$ 0.42\\
        
        Coreset & 77.70 $\pm$ 1.31 & 82.78 $\pm$ 0.90 & 83.79 $\pm$ 0.74\\
        DBAL & 79.28 $\pm$ 0.78 & 81.16 $\pm$ 0.83 & \textbf{85.58} $\pm$ \textbf{0.19} \\
        BALD & 78.67 $\pm$ 0.39 & \textbf{82.95} $\pm$ \textbf{0.43} & 84.11 $\pm$ 0.30\\
        VAAL & 76.50 $\pm$ 0.70 & 79.12 $\pm$ 0.62 & 82.86 $\pm$ 0.69\\
        QBC & 78.22 $\pm$ 1.84 & 82.68 $\pm$ 0.54 & 85.34 $\pm$ 1.26\\
        UC & \textbf{80.03} $\pm$ \textbf{0.27} & 79.49 $\pm$ 0.37 & 85.45 $\pm$ 0.69 \\
        \bottomrule
        \end{tabular}
        }
        \caption{CIFAR10 Test Accuracy on L$_{1}^{0}$.The base model accuracy is $68.02$.}
        \label{tab:cif10_l0_1_supp}
        
    \end{minipage}

        \end{tabular}

    \end{center}
\end{table}

\begin{table}[ht]
    \begin{center}
        \begin{tabular}{c c}
            
        \begin{minipage}{0.5\linewidth}   
        \centering
        \resizebox{0.85\linewidth}{!}{
        
        \begin{tabular}{l|cccc}
        \toprule
        Methods & $20\%$ & $30\%$ & $40\%$ \\
        \midrule
        RSB & 77.02 $\pm$ 0.71 & 80.50 $\pm$ 0.30 & 83.82 $\pm$ 0.37\\
        
        Coreset & 74.67 $\pm$ 0.82 & 81.14 $\pm$ 0.92 & 81.58 $\pm$ 1.19\\
        DBAL & 75.9 $\pm$ 0.25 & 80.58 $\pm$ 3.16 & 83.75 $\pm$ 0.88 \\
        BALD & 76.19 $\pm$ 0.86 & \textbf{83.26} $\pm$ \textbf{0.36} & \textbf{85.39} $\pm$ \textbf{0.97}\\
        VAAL & 76.88 $\pm$ 0.96 & 81.30 $\pm$ 0.29 & 82.63 $\pm$ 0.55\\
        QBC & \textbf{78.38} $\pm$ \textbf{0.79} & 81.39 $\pm$ 3.3 & 85.16 $\pm$ 0.77\\
        UC & 78.16 $\pm$ 0.85 & 81.80 $\pm$ 0.45 & 84.91 $\pm$ 0.69 \\
        
        \bottomrule
        \end{tabular}}
        
        \caption{CIFAR10 Test Accuracy on L$_{2}^{0}$. The base model accuracy is $70.34$.}
        \label{tab:CIFAR10_exp_budget_10k}
        
    \end{minipage} 
    
    &
    
    \begin{minipage}{0.5\linewidth}
    \centering
    \resizebox{0.85\linewidth}{!}{
        \begin{tabular}{l|cccc}
        \toprule
        Methods & $20\%$ & $30\%$ & $40\%$ \\
        \midrule
        RSB & 75.84 $\pm$ 1.91 & 80.93 $\pm$ 1.20 & 83.17 $\pm$ 0.52\\
        
        Coreset & 79.42 $\pm$ 0.47 & 81.62 $\pm$ 0.86 & 83.82 $\pm$ 0.18\\
        DBAL & \textbf{79.48} $\pm$ \textbf{0.35} & 82.27 $\pm$ 1.23 & 84.74 $\pm$ 0.14 \\
        BALD & 77.58 $\pm$ 0.88 & 82.11 $\pm$ 0.65 & 84.58 $\pm$ 0.42\\
        VAAL & 77.45 $\pm$ 1.21 & 79.38 $\pm$ 1.08 & 82.90 $\pm$ 0.94\\
        QBC & 78.60 $\pm$ 0.43 & \textbf{82.76} $\pm$ \textbf{0.92} & \textbf{85.54} $\pm$ \textbf{0.69}\\
        UC & 76.97 $\pm$ 0.79 & 81.35 $\pm$ 0.82 & 84.65 $\pm$ 0.30 \\
        \bottomrule
        \end{tabular}}
        \caption{CIFAR10 Test Accuracy on L$_{3}^{0}$.The base model accuracy is $68.19$.}
        \label{tab:CIFAR10_exp_budget_10k_1}
        
    \end{minipage}

        \end{tabular}
        
    \end{center}
\end{table}

\begin{table}[ht]
    \begin{center}
        \begin{tabular}{c c}
            
        \begin{minipage}{0.5\linewidth}   
        \centering
        \resizebox{0.85\linewidth}{!}{
        
        \begin{tabular}{l|cccc}
        \toprule
        Methods & $20\%$ & $30\%$ & $40\%$ \\
        \midrule
        RSB & 78.59 $\pm$ 0.91 & 81.81 $\pm$ 0.71 & 83.46 $\pm$ 0.18\\
        
        Coreset & 77.17 $\pm$ 1.82 & 81.37 $\pm$ 0.41 & 83.13 $\pm$ 1.54\\
        DBAL & 75.87 $\pm$ 0.61 & 83.00 $\pm$ 0.79 & 85.13 $\pm$ 1.25 \\
        BALD & 78.49 $\pm$ 0.46 & 83.21 $\pm$ 0.66 & 85.06 $\pm$ 0.60\\
        VAAL & 73.67 $\pm$ 1.47 & 79.49 $\pm$ 1.27 & 82.98 $\pm$ 0.78\\
        QBC & \textbf{78.61} $\pm$ \textbf{1.65} & \textbf{83.81} $\pm$ \textbf{0.49} & 85.35 $\pm$ 0.82\\
        UC & 77.38 $\pm$ 1.17 & 81.82 $\pm$ 1.86 & \textbf{85.62} $\pm$ \textbf{0.30} \\
        \bottomrule
        
        \end{tabular}

        }

        \caption{CIFAR10 Test Accuracy on L$_{4}^{0}$. The base model accuracy is $67.19$.}
        \label{tab:CIFAR10_tab11_supp}
        
    \end{minipage} 
    
    &
    
    \begin{minipage}{0.5\linewidth}
    \centering
    \resizebox{0.85\linewidth}{!}{
        \begin{tabular}{l|cccc}
        \toprule
        Methods & $20\%$ & $30\%$ & $40\%$ \\
        \midrule
        RSB & 46.67 $\pm$ 0.30 & 51.43 $\pm$ 0.81 & 55.06 $\pm$ 0.35\\
        
        Coreset & \textbf{47.33} $\pm$ \textbf{0.64} & 49.73 $\pm$ 0.92 & 57.05 $\pm$ 0.40\\
        DBAL & 45.53 $\pm$ 2.33 & 51.04 $\pm$ 0.49 & \textbf{58.06} $\pm$ \textbf{0.51} \\
        BALD & 47.10 $\pm$ 1.24 & 50.40 $\pm$ 0.88 & 55.65 $\pm$ 0.34\\
        VAAL & 39.73 $\pm$ 0.43 & 50.95 $\pm$ 0.88 & 55.23 $\pm$ 0.63\\
        QBC & 46.04 $\pm$ 0.57 & \textbf{53.20} $\pm$ \textbf{0.38} & 57.63 $\pm$ 0.49\\
        UC & 41.37 $\pm$ 1.29 & 52.97 $\pm$ 0.83 & 55.45 $\pm$ 0.62 \\
        
        \bottomrule
        \end{tabular}}
        
        \caption{CIFAR100 Test Accuracy on L$_{0}^{0}$. The base model accuracy is $34.73$.}
        
        \label{tab:CIFAR100_tab12_supp}
        
    \end{minipage}

        \end{tabular}
        
    \end{center}
\end{table}

\begin{table}[ht]
    \begin{center}
        \begin{tabular}{c c}
        
    \begin{minipage}{0.5\linewidth}
    \centering
    \resizebox{0.85\linewidth}{!}{

        \begin{tabular}{l|cccc}
        \toprule
        Methods & $20\%$ & $30\%$ & $40\%$ \\
        \midrule
        RSB & 45.58 $\pm$ 0.19 & \textbf{53.45} $\pm$ \textbf{0.28} & 56.98 $\pm$ 0.31\\
        
        Coreset & \textbf{46.05} $\pm$ \textbf{0.46} & 52.04 $\pm$ 0.23 & 58.11 $\pm$ 0.12\\
        DBAL & 41.32 $\pm$ 0.23 & 52.16 $\pm$ 0.81 & 58.00 $\pm$ 0.68 \\
        BALD & 43.57 $\pm$ 0.80 & 53.27 $\pm$ 0.12 & 56.87 $\pm$ 0.73\\
        VAAL & 42.70 $\pm$ 0.75 & 48.86 $\pm$ 1.61 & 54.81 $\pm$ 1.23\\
        QBC & 45.61 $\pm$ 0.74 & 53.31 $\pm$ 0.91 & \textbf{58.21} $\pm$ \textbf{0.22}\\
        UC & 37.48 $\pm$ 0.45 & 53.01 $\pm$ 0.16 & 57.80 $\pm$ 0.09 \\
        
        \bottomrule
        \end{tabular}}
        \caption{CIFAR100 Test Accuracy on L$_{1}^{0}$.The base model accuracy is $32.73$.}
        \label{tab:CIFAR10_exp_budget_10k_3}
        
    \end{minipage}
    
    & 
    
    \begin{minipage}{0.5\linewidth}   
        \centering
        \resizebox{0.85\linewidth}{!}{
        
        \begin{tabular}{l|cccc}
        \toprule
        Methods & $20\%$ & $30\%$ & $40\%$ \\
        \midrule
        RSB & 44.71 $\pm$ 0.64 & 50.01 $\pm$ 0.36 & 56.27 $\pm$ 0.84\\
        
        Coreset & 46.00 $\pm$ 0.79 & 53.48 $\pm$ 0.61 & 57.22 $\pm$ 0.69\\
        DBAL & 44.06 $\pm$ 0.39 & 49.29 $\pm$ 1.00 & 57.40 $\pm$ 0.34 \\
        BALD & \textbf{46.78} $\pm$ \textbf{0.52} & 52.34 $\pm$ 0.90 & 54.97 $\pm$ 0.98\\
        VAAL & 44.75 $\pm$ 0.57 & 49.72 $\pm$ 0.40 & 55.77 $\pm$ 0.62\\
        QBC & 46.20 $\pm$ 0.72 & 53.15 $\pm$ 0.90 & \textbf{57.96} $\pm$ \textbf{0.65}\\
        UC & 43.94 $\pm$ 0.60 & \textbf{53.75} $\pm$ \textbf{0.50} & 55.10 $\pm$ 0.95 \\
        
        \bottomrule
        \end{tabular}}
        
        \caption{CIFAR100 Test Accuracy on L$_{2}^{0}$. The base model accuracy is $34.66$.}
        \label{tab:CIFAR10_exp_budget_10k_4}
        
    \end{minipage}

        \end{tabular}
        
    \end{center}
\end{table}

\begin{table}[ht]
    \begin{center}
        \begin{tabular}{c c}
        
        \begin{minipage}{0.5\linewidth}
    \centering
    \resizebox{0.85\linewidth}{!}{
        \begin{tabular}{l|cccc}
        \toprule
        Methods & $20\%$ & $30\%$ & $40\%$ \\
        \midrule
        RSB & 42.46 $\pm$ 0.44 & 52.66 $\pm$ 0.66 & 54.15 $\pm$ 0.43\\
        
        Coreset & 45.98 $\pm$ 0.83 & \textbf{54.34} $\pm$ \textbf{0.53} & 56.96 $\pm$ 0.95\\
        DBAL & 45.49 $\pm$ 0.51 & 48.84 $\pm$ 0.42 & 57.65 $\pm$ 0.46 \\
        BALD & \textbf{47.21} $\pm$ \textbf{1.26} & 52.53 $\pm$ 0.42 & 55.39 $\pm$ 0.72\\
        VAAL & 44.93 $\pm$ 1.61 & 46.27 $\pm$ 0.72 & 56.65 $\pm$ 0.60\\
        QBC & 46.50 $\pm$ 0.56 & 53.49 $\pm$ 0.53 & \textbf{57.68} $\pm$ \textbf{0.51}\\
        UC & 46.96 $\pm$ 0.41 & 53.07 $\pm$ 0.57 & 56.35 $\pm$ 0.79 \\
        \bottomrule
        \end{tabular}}
        \caption{CIFAR100 Test Accuracy on L$_{3}^{0}$.The base model accuracy is $30.44$.}
        \label{tab:CIFAR10_exp_budget_10k_6}
        
    \end{minipage}
    
    & 
    
            \begin{minipage}{0.5\linewidth}   
        \centering
        \resizebox{0.85\linewidth}{!}{
        
        \begin{tabular}{l|cccc}
        \toprule
        Methods & $20\%$ & $30\%$ & $40\%$ \\
        \midrule
        RSB & 41.15 $\pm$ 0.89 & 50.61 $\pm$ 0.40 & 56.77 $\pm$ 0.55\\
        
        Coreset & 45.72 $\pm$ 0.77 & 52.22 $\pm$ 0.54 & 56.28 $\pm$ 0.45\\
        DBAL & 44.71 $\pm$ 0.57 & 52.33 $\pm$ 0.49 & 56.52 $\pm$ 0.51 \\
        BALD & 40.35 $\pm$ 0.75 & 51.87 $\pm$ 0.60 & 57.40 $\pm$ 0.40\\
        VAAL & 44.86 $\pm$ 1.69 & 51.32 $\pm$ 1.54 & 53.82 $\pm$ 1.08\\
        QBC & \textbf{45.93} $\pm$ \textbf{0.46} & \textbf{53.12} $\pm$ \textbf{0.55} & \textbf{57.78} $\pm$ \textbf{0.49}\\
        UC & 43.07 $\pm$ 0.74 & 49.89 $\pm$ 0.79 & 56.15 $\pm$ 0.52 \\
        \bottomrule
        \end{tabular}}
        
        \caption{CIFAR100 Test Accuracy on L$_{4}^{0}$.The base model accuracy is $34.85$.}
        
        \label{tab:CIFAR100_tab16_supp}
        
    \end{minipage}

        \end{tabular}
        
    \end{center}
\end{table}

\end{document}